\pdfoutput=1

\documentclass[11pt]{article}

\usepackage[preprint]{acl}

\usepackage{times}
\usepackage{latexsym}
\usepackage[T1]{fontenc}
\usepackage[utf8]{inputenc}
\usepackage{microtype}
\usepackage{inconsolata}
\usepackage{graphicx}
\usepackage{amsmath}
\usepackage{booktabs}
\usepackage{hyperref}
\usepackage{tabularx}
\usepackage{siunitx}
\usepackage{multirow}
\usepackage{amsfonts}
\usepackage{amsmath}
\usepackage{subcaption}
\usepackage{float}
\usepackage{makecell}

\title{Chronocept: Instilling a Sense of Time in Machines}

\author{
    Krish Goel\thanks{Equal contribution.}, 
    Sanskar Pandey\footnotemark[1], 
    KS Mahadevan, 
    Harsh Kumar, 
    Vishesh Khadaria \\
    \texttt{krish@projectendgame.tech, pandeysanskar854@gmail.com, mahadevanks26@gmail.com,}
    \\
    \texttt{kumarharsh3014@gmail.com, khadariavishesh@gmail.com}
}

\begin{document}
\maketitle

\begin{abstract}
Human cognition is deeply intertwined with a sense of time, known as \textit{Chronoception}. This sense allows us to judge how long facts remain valid and when knowledge becomes outdated. Despite progress in vision, language, and motor control, AI still struggles to reason about temporal validity. We introduce Chronocept, the first benchmark to model temporal validity as a continuous probability distribution over time. Using skew-normal curves fitted along semantically decomposed temporal axes, Chronocept captures nuanced patterns of emergence, decay, and peak relevance. It includes two datasets: Benchmark I (atomic facts) and Benchmark II (multi-sentence passages). Annotations show strong inter-annotator agreement (84\% and 89\%). Our baselines predict curve parameters - location, scale, and skewness - enabling interpretable, generalizable learning and outperforming classification-based approaches. Chronocept fills a foundational gap in AI's temporal reasoning, supporting applications in knowledge grounding, fact-checking, retrieval-augmented generation (RAG), and proactive agents. Code and data are publicly available.
\end{abstract}

\section{Introduction}
\label{sec:introduction}
Humans effortlessly track how information changes in relevance over time. We instinctively know when facts emerge, become useful, or fade into obsolescence - a cognitive ability known as Chronoception \citep{Fontes2016-fk, Zhou2019-hk}. This higher-order perception of time plays a crucial role in how we evaluate the persistence and usefulness of information in real-world contexts. Despite excelling in pattern recognition \citep{he2016deep}, language generation \citep{brown2020language}, and motor control \citep{levine2016end}, modern AI systems remain largely insensitive to the temporal validity of the information they process.

Prior work has advanced temporal understanding via event ordering \citep{Allen1983-fa, Ning2020-xw, Wen2021-cl}, timestamp prediction \citep{Kanhabua2008-ve, Kumar2012-ea, Das2017-ex}, and temporal commonsense reasoning \citep{Zhou2019-hk}. However, these approaches often reduce time to static labels or binary transitions. Even recent efforts in temporal validity change prediction \citep{wenzel-jatowt-2024-temporal} model shifts as discrete class changes, neglecting the gradual and asymmetric nature of temporal decay.

We introduce Chronocept, a benchmark that models temporal validity as a continuous probability distribution over time. Using a skewed-normal distribution over logarithmic time, parameterized by location (\(\xi\)), scale (\(\omega\)), and skewness (\(\alpha\)) \citep{Azzalini1986-gp, Schmidt2017-pf}, Chronocept captures subtle temporal patterns such as delayed peaks and asymmetric decay.

To support structured supervision, we decompose each sample along semantic temporal axes. We release two benchmarks: Benchmark I features atomic factual statements, and Benchmark II contains multi-sentence passages with temporally interdependent elements. High inter-annotator agreement across segmentation, axis labeling, and curve parameters validates annotation quality.

We benchmark a diverse set of models, including linear regression, SVMs, XGBoost, FFNNs, Bi-LSTMs, and BERT \citep{devlin2019bert}. FFNNs perform best on the simpler Benchmark I, while Bi-LSTMs excel on the more complex Benchmark II. Surprisingly, fine-tuned BERTs do not outperform simpler architectures. To assess the role of temporal structure, we conduct ablations that remove or shuffle temporal axes during training - both lead to marked performance drops. 

Chronocept enables several downstream applications. In Retrieval-Augmented Generation (RAG), temporal curves guide time-sensitive retrieval; in fact-checking, they help flag decaying or stale facts. Most importantly, Chronocept lays the foundation for proactive AI systems that reason not just about what to do, but when to do it \citep{miksik2020building}.

All resources - dataset, and baseline implementations - are publicly available to support future research in machine time-awareness.

\section{Related Work}
\label{sec:related_work}
\subsection{Temporal Validity Prediction}
In the earliest attempt to formalize the temporal validity of information, \citet{Takemura2012-nv} proposed the concept of ``content viability'' by classifying tweets into ``read now,'' ``read later,'' and ``expired'' categories, to prioritize timeliness in information consumption. However, their approach assumed a rigid, monotonic decay of relevance, failing to model scenarios where validity peaks later rather than at publication. This restricted its applicability beyond real-time contexts such as Twitter streams.

\citet{Almquist2019-un} extended this work by defining a ``validity period'' and effectively proposing a ``content expiry date'' for sentences, using linguistic and statistical features. However, their reliance on static time classes (e.g., hours, days, weeks) sacrificed granularity, and their approach required explicit feature engineering rather than leveraging more advanced, data-driven methods \citep{Das2017-ex}.

Traditional approaches \citep{Almquist2019-un, Lynden2023-lj, Hosokawa2023-ay} mostly treat validity as binary, where information is either valid or invalid at a given time, this can be formulated as:

\begin{equation}
    \label{eq:tv}
    \small
    \text{validity}_i(t) =
    \begin{cases}
        \text{True} & \text{if information $i$ is valid at $t$}, \\
        \text{False} & \text{otherwise}
    \end{cases}
\end{equation}

where $i$ represents the information under consideration and $t$ denotes the time at which its validity is evaluated. However, this model overlooks gradual decay, recurrence, and asymmetric relevance patterns.

More recently, \citet{wenzel-jatowt-2024-temporal} introduced Temporal Validity Change Prediction (TVCP), which models how context alters a statement’s validity window. However, it does not quantify validity as a continuous probability distribution over time.

Chronocept advances this field by defining temporal validity as a continuous probability distribution, allowing a more precise and flexible representation of how information relevance evolves.

\subsection{Temporal Reasoning and Commonsense}
Temporal reasoning has largely focused on event ordering \citep{Allen1983-fa, Wen2021-cl, Ning2020-xw}, predicting temporal context \citep{Kanhabua2008-ve, Kumar2012-ea, Das2017-ex, Luu2021-jh, Jatowt2013-up}, and commonsense knowledge \citep{Zhou2019-hk}. While these studies laid the groundwork for understanding event sequences, durations, and frequencies, recent work has expanded into implicit or commonsense dimensions of temporal reasoning.

TORQUE (Ning et al., 2020) is a benchmark designed for answering temporal ordering questions, while TRACIE, along with its associated model SYMTIME (Zhou et al., 2021), primarily ensures temporal-logical consistency rather than modeling truth probabilities.

McTACO \citep{Zhou2019-hk} evaluates temporal commonsense across five dimensions: event duration, ordering, frequency, stationarity, and typical time of occurrence. McTACO assesses whether a given statement aligns with general commonsense expectations, and does not quantify the likelihood of a statement being true over time.

Recent work \citealp{Wenzel2023-ta, Jain2023-ww} has explored how LLMs handle temporal commonsense, exposing inconsistencies in event sequencing and continuity. However, these studies do not incorporate probabilistic modeling of temporal validity - a core focus of Chronocept, which models truthfulness as a dynamic, evolving probability distribution.

\subsection{Dataset Structuring for Temporal Benchmarks}
Temporal annotation frameworks like TimeML \citep{Pustejovsky2003-od} and ISO-TimeML \citep{Pustejovsky2010-lf} focus on static event relationships, often suffering from low inter-annotator agreement due to event duration ambiguities. The TimeBank series \citep{Pustejovsky2003-oz, Cassidy2014-qt} and TempEval challenges \citep{Verhagen2007-wz, Verhagen2010-bl, UzZaman2012-rf} expanded evaluations but remained limited in modeling evolving event validity.

In response, \citet{Ning2018-wk} proposed a multi-axis annotation scheme that structures events into eight distinct categories - Main, Intention, Opinion, Hypothetical, Negation, Generic, Static, and Recurrent. Additionally, the scheme prioritizes event start points over full event intervals, reducing ambiguity and significantly improving IAA scores. Chronocept builds on this by refining multi-axis annotation to model temporal validity, capturing how information relevance shifts over time through probabilistic distributions.

\subsection{Statistical Modeling of Temporal Data Using Skewed Normal Distribution}
Traditional normal distributions often fail to capture skewed temporal patterns. The skew-normal distribution \citep{Azzalini1986-gp, Azzalini1996-xf} provides a more flexible alternative by incorporating asymmetry, improving accuracy in modeling time-dependent information relevance \citep{Schmidt2017-pf}. Chronocept employs this distribution to capture various temporal behaviors, including gradual decay, peak relevance, and rapid obsolescence.

\section{Chronocept: Task \& Benchmark Design}
\label{sec:task_and_benchmark_design}
\subsection{Problem Definition}
Temporal Validity Prediction (TVP) of Information seeks to model how long a factual statement remains true after it is published.

We formalize Temporal Validity Prediction as a probabilistic task of modeling information's relevance as a continuous probability distribution over time, rather than the binary‐or‐multiclass settings common in earlier work.

Let \(T \subseteq \mathbb{R}_{\geq 0}\) denote the time domain, where \(t \geq 0\) represents the elapsed time since publication of information \(i\).

Then, we define a binary random variable,
\begin{equation}
    \label{eq:validity_binary_rv}
    \text{validity}_i(t) \in \{0,1\}
\end{equation}

where \(\text{validity}_i(t)=1\) indicates that the information \(i\) is valid at time \(t\), and \(\text{validity}_i(t)=0\) otherwise.

Rather than predicting \(\text{validity}_i(t)\) directly, TVP aims to learn a continuous probability density function \(p_i(t)\)
\begin{equation}
    \label{eq:validity_pdf}
    p_i(t) = P\bigl(\text{validity}_i(t)=1\bigr),
    \:
    p_i : T \to [0,1]
\end{equation}

Accordingly, the probability that the statement remains valid throughout any
interval \([a,b]\subseteq T\) is given by
\begin{equation}
    \label{eq:integral}
    P\Bigl(\forall\, t \in [a,b], \ \text{validity}_i(t)=1\Bigr) = \int_a^b p_i(t) \, dt
\end{equation}

Crucially, the model does not impose rigid boundary constraints - such as
\(p_i(0)=1\) or monotonic decay - thereby permitting the learned distribution to capture complex temporal phenomena, including delayed onset, non-monotonic plateaus, and intermittent resurgences \citep{Takemura2012-nv, Almquist2019-un}

\subsection{Modeling Temporal Validity}
We model the temporal validity of statements using a probability curve, with likelihood of being valid on the Y-axis and time since publication on the X-axis. To reduce ambiguity, sentences are decomposed along semantically distinct axes. A skew-normal distribution on a logarithmic time scale captures the validity dynamics.

\paragraph{Axes-Based Decomposition.}
We adopt the multi‐axis annotation scheme of \citet{Ning2018-wk} (MATRES), which partitions each sentence into eight semantically coherent axes (Main, Intention, Opinion, Hypothetical, Generic, Negation, Static, Recurrent). By isolating relation annotation within each axis, MATRES reduces cross-category ambiguity and better aligns with human temporal perception.

In our ablation \autoref{appendix:multi_axis_study}, removing axis features increases MSE by 4.57\%, confirming that axis‐level signals are essential for precise temporal modeling.\\

\paragraph{Skewed Normal Distribution.}
We model temporal validity using the skewed normal distribution, a generalization of the Gaussian with a shape parameter \(\alpha\) that captures asymmetry. This enables representation of non-symmetric temporal patterns such as delayed onset, gradual decay, or skewed relevance, which symmetric (Gaussian) or memoryless (exponential) distributions fail to model.

The probability density function is:
\begin{equation}\label{eq:skewnormal}
    f(x; \xi, \omega, \alpha) = \frac{2}{\omega} \, \phi\left(\frac{x-\xi}{\omega}\right) \, \Phi\left(\alpha\, \frac{x-\xi}{\omega}\right)
\end{equation}
where:
\begin{itemize}
    \setlength{\itemsep}{0pt}
    \item \(\phi(z)\) is the standard normal PDF,
    \item \(\Phi(z)\) is the standard normal CDF,
    \item \(\xi\) is the location parameter - determining the time at which an event is most likely valid,
    \item \(\omega\) is the scale parameter - governing the duration of validity, and
    \item \(\alpha\) is the shape parameter - controlling skewness (with positive values yielding right skew and negative values left skew).
\end{itemize}

Quantitative comparisons against Gaussian, log-normal, exponential and gamma distributions in \autoref{appendix:curve_fitting} support this choice.\\

\paragraph{Logarithmic Time Scale.}
Linear time yields sparse coverage over key intervals, particularly at minute-level granularity. To address this, we compress the time axis using a monotonic logarithmic transformation:
\begin{equation}
    \label{eq:time_transformation}
    t' = \log_{1.1}(t)
\end{equation}

We default to a base of \(1.1\) for the near-linear spacing across canonical intervals (e.g., minutes, days, decades) while preserving granularity. Chronocept’s target values remain compatible with alternative bases. See \autoref{appendix:log_conversion} for the base transformation framework, compression analysis, and the provided code implementation.

\section{Dataset Creation}
\label{sec:dataset}
\subsection{Benchmark Generation \& Pre-Filtering}
Chronocept comprises two benchmarks to facilitate evaluation across varying complexity levels. Benchmark I consists of 1,254 samples featuring simple, single-sentence texts with clear temporal relations - ideal for baseline reasoning - while Benchmark II includes 524 samples with complex, multi-sentence texts capturing nuanced, interdependent temporal phenomena.

Synthetic samples were generated using the GPT-o1\footnote{\url{https://openai.com/o1}} model \citep{openai2024o1} with tailored prompts to ensure temporal diversity across benchmarks. Full prompts for both benchmarks are disclosed in \autoref{appendix:synthetic} for reproducibility. No real-world or personally-identifying data was used, ensuring complete privacy.

In the pre-annotation phase, SBERT\footnote{\texttt{all-MiniLM-L6-v2} available at \url{https://huggingface.co/sentence-transformers/all-MiniLM-L6-v2}} \citep{Reimers2019-hb} and TF-IDF embeddings were generated for all samples, and pairwise cosine similarities were calculated. Samples with SBERT or TF-IDF similarity exceeding 0.7 (70\%) were removed to reduce semantic and lexical redundancy.

Annotation guidelines are disclosed in \autoref{appendix:annotation_guidelines} and were continuously accessible during annotation.

\subsection{Annotation Workflow}

\paragraph{Annotation Process.}
Our protocol consists of three steps: (i) \emph{Temporal Segmentation} – partitioning text into coherent subtexts that preserve temporal markers; (ii) \emph{Axis Categorization} – assigning each segment to one of eight temporal axes (Main, Intention, Opinion, Hypothetical, Generic, Negation, Static, Recurrent); and (iii) \emph{Temporal Validity Distribution Plotting} – annotating a skewed normal distribution, parameterized by location (\(\xi\)), scale (\(\omega\)), and skewness (\(\alpha\)), over a logarithmic time axis.

To ensure interpretability and consistency, all parent texts are written in the present tense, distributions are anchored at \(t=0\), and multimodal curves are excluded. Additionally, any samples that did not exhibit a clearly assignable main timeline or violated these constraints were flagged and discarded during the annotation process.

\subsection{Annotator Training \& Quality Control}
Eight third-year B.Tech. students with relevant coursework in Natural Language Processing, Machine Learning, and Information Retrieval participated. They underwent a 1-hour training session and a supervised warm-up on 50 samples. Agreement thresholds were set at ICC > 0.90 for numerical annotations, Jaccard Index > 0.75 for segment-level annotations, and $P_k$ < 0.15 for segmentation consistency during this warm-up phase.

Each sample was annotated independently by two annotators. Quality control included daily reviews of 10\% of annotations, a limit of 70 samples per annotator per day to mitigate fatigue, and automated flagging of samples with segmentation mismatches, target deviations >2\(\sigma\), or P\(_k\) > 0.2. Discrepancies were adjudicated or, if unresolved, discarded.

No personal or identifying information was collected or stored during the annotation process.

\paragraph{Handling Edge Cases and Final Resolution.}
Ambiguous samples were flagged or discarded following the three-phase filtering scheme. For segmentation and axis labels, a union-based approach retained all plausible interpretations, recognizing that axis confusion may encode aspects of human temporal cognition useful for future modeling. For temporal validity targets (\(\xi\), \(\omega\), \(\alpha\)), annotator values were averaged to yield smooth probabilistic supervision rather than discrete target selection.\\

\begin{figure*}[h]
    \centering
    \includegraphics[width=1\linewidth]{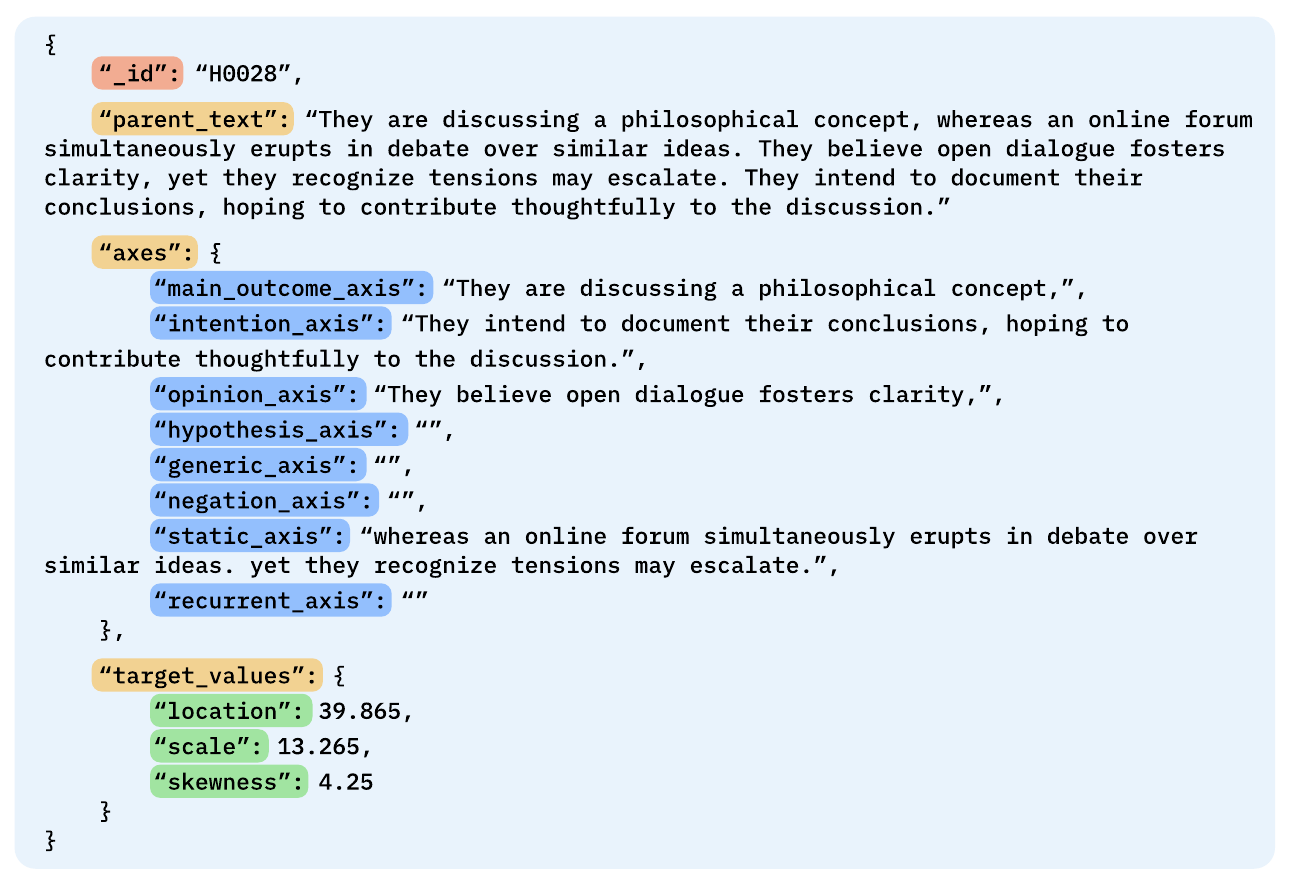}
    \caption{Composition of samples in Chronocept benchmarks.}
    \label{fig:sample_composition}
\end{figure*}

\subsection{Inter-Annotator Agreement (IAA)}
We evaluate Inter-Annotator Agreement (IAA) using stage-specific metrics aligned with each step of the annotation task. Segmentation quality is assessed using the $P_k$ metric \citep{beeferman-etal-1997-text}, axis categorization consistency is measured using the Jaccard Index, and agreement on the final temporal validity parameters (\(\xi\), \(\omega\), \(\alpha\)) is quantified using the Intraclass Correlation Coefficient (ICC).

We report only ICC as the benchmark-wide IAA, refraining from aggregating agreement across stages, as segmentation and axis categorization, while enriching the dataset structure, do not directly impact the core prediction task, which depends solely on the parent text and its annotated temporal validity distribution.

Agreement statistics across both benchmarks are summarized in \autoref{tab:iaa_results}. We observed notable confusion between the \textit{Generic} and \textit{Static} axes during the early stages of annotation, particularly in the warm-up phase. This source of disagreement is analyzed in detail in \autoref{appendix:generic_static_confusion_analysis}.

\begin{table}[h]
\centering
\begin{tabular}{lcc}
\toprule
\textbf{IAA Metric} & \textbf{BI} & \textbf{BII} \\
\midrule
\textbf{ICC} & 0.843 & 0.893\\
\textbf{Jaccard Index} & 0.624 & 0.731 \\
\textbf{$P_k$ Metric} & 0.233 & 0.009 \\
\bottomrule
\end{tabular}
\caption{IAA metrics for segmentation, axis categorization, and temporal validity annotation across both benchmarks. For $P_k$, lower is better, with values ranging from 0 (perfect agreement) to 1 (chance-level).}
\label{tab:iaa_results}
\end{table}

\subsection{Dataset Design}
Each Chronocept sample captures the temporal dynamics of factual information through a structured annotation format, as illustrated in \autoref{fig:sample_composition}.

\vspace{0.5em}
\noindent
\paragraph{Parent Text.} A single sentence serving as the basis for annotation.

\vspace{0.5em}
\noindent
\paragraph{Temporal Axes.} Each parent text is segmented into subtexts annotated along eight temporal axes:

\begin{itemize}
    \setlength{\itemsep}{0pt}
    \item \textbf{Main:} Core verifiable events.
    \item \textbf{Intention:} Future plans or desires.
    \item \textbf{Opinion:} Subjective viewpoints.
    \item \textbf{Hypothetical:} Conditional or imagined events.
    \item \textbf{Negation:} Denied or unfulfilled events.
    \item \textbf{Generic:} Timeless truths or habitual patterns.
    \item \textbf{Static:} Unchanging states in context.
    \item \textbf{Recurrent:} Repeated temporal patterns.
\end{itemize}

\noindent
\paragraph{Target Values.} Temporal validity is quantified by three parameters:
\begin{itemize}
    \setlength{\itemsep}{0pt}
    \item \textbf{\(\xi\) (Location):} The time point of peak validity.
    \item \textbf{\(\omega\)} (Scale): The duration over which validity is maintained.
    \item \textbf{\(\alpha\) (Skewness):} The asymmetry of the validity curve.
\end{itemize}

\subsection{Dataset Statistics \& Splits}
Stratified sampling over the axes distribution was applied to partition the datasets into training (70\%), validation (20\%), and test (10\%) splits, ensuring equitable axis coverage. \autoref{tab:dataset} summarizes the splits for both benchmarks. The axes distribution, calculated based on non-null annotations for each sample, is detailed in \autoref{tab:axis_distribution}.

\begin{table}[h]
    \centering
    \sisetup{group-separator={,}}
    \begin{tabular}{lccc}
        \toprule
        \textbf{Benchmark} & \textbf{Training} & \textbf{Validation} & \textbf{Test} \\
        \midrule
        Benchmark I & 878 & 247 & 129 \\
        Benchmark II & 365 & 104 & 55 \\
        \bottomrule
    \end{tabular}
    \caption{Dataset Composition and Splits.}
    \label{tab:dataset}
\end{table}

\begin{table}[h]
    \fontsize{10pt}{12pt}\selectfont
    \centering
    \begin{tabular}{lcc}
        \toprule
        \textbf{Temporal Axis} & \textbf{Benchmark I} & \textbf{Benchmark II} \\
        \midrule
        Main Axis & 1254 & 524 \\
        Static Axis & 516 & 513 \\
        Generic Axis & 228 & 116 \\
        Hypothetical Axis & 136 & 182 \\
        Negation Axis & 240 & 200 \\
        Intention Axis & 165 & 522 \\
        Opinion Axis & 328 & 519 \\
        Recurrent Axis & 348 & 198 \\
        \bottomrule
    \end{tabular}
    \caption{Distribution of annotated temporal axes across Benchmark I and Benchmark II.}
    \label{tab:axis_distribution}
\end{table}

Token-level\footnote{Tokenization performed using SpaCy's \texttt{en\_core\_web\_sm} model: ~\url{https://spacy.io/api/tokenizer}} and target parameter-level statistics for both benchmarks are summarized in \autoref{tab:sentence_stats} and \autoref{tab:temporal_params}.

\begin{table}[h]
    \centering
    \begin{tabular}{lcc}
        \toprule
        \textbf{Benchmark} & \textbf{Mean Length (\(\mu\))} & \textbf{SD (\(\sigma\))} \\
        \midrule
        Benchmark I & 16.41 tokens & 1.56 tokens \\
        Benchmark II & 56.21 tokens & 6.21 tokens \\
        \bottomrule
    \end{tabular}
    \caption{Sentence Length Statistics for Benchmarks.}
    \label{tab:sentence_stats}
\end{table}

\begin{table*}[h]
    \centering
    \begin{tabular}{lcccccc}
        \toprule
        \textbf{Parameter} & \multicolumn{2}{c}{\textbf{Location (\(\xi\))}} & \multicolumn{2}{c}{\textbf{Duration (\(\omega\))}} & \multicolumn{2}{c}{\textbf{Skewness (\(\alpha\))}} \\
        \cmidrule(lr){2-3} \cmidrule(lr){4-5} \cmidrule(lr){6-7}
        \textbf{Benchmark} & \textbf{Mean (\(\mu\))} & \textbf{SD (\(\sigma\))} & \textbf{Mean (\(\mu\))} & \textbf{SD (\(\sigma\))} & \textbf{Mean (\(\mu\))} & \textbf{SD (\(\sigma\))} \\
        \midrule
        Benchmark I & 54.2803 & 20.4169 & 11.5474 & 3.7725 & -0.0158 & 1.3858 \\
        Benchmark II & 46.1511 & 13.3839 & 9.5553 & 2.5725 & 0.0275 & 1.1773 \\
        \bottomrule
    \end{tabular}
    \caption{Temporal Parameter Distribution Statistics for Benchmarks.}
    \label{tab:temporal_params}
\end{table*}

\subsection{Accessibility and Licensing}
The Chronocept dataset is released under the Creative Commons Attribution 4.0 International (CC-BY 4.0)\footnote{\url{https://creativecommons.org/licenses/by/4.0}} license, allowing unrestricted use with proper attribution. It is publicly available on Hugging Face Datasets at: \url{https://huggingface.co/datasets/krishgoel/chronocept}.

\section{Baseline Model Performance}
\label{sec:baseline_model_performance}
\subsection{Task Scope and Evaluation Focus}
Chronocept models temporal validity as a structured regression task over low-dimensional parameters: location (\(\xi\)), scale (\(\omega\)), and skewness (\(\alpha\)), predicted from annotated parent texts. Unlike prior work on event ordering \citep{Pustejovsky2003-oz}, commonsense classification \citep{Zhou2019-hk}, or temporal shift detection \citep{wenzel-jatowt-2024-temporal}, segmentation and axis labels are treated as preprocessing and not modeled at inference.

Evaluation spans three dimensions: regression accuracy (MSE, MAE, $R^2$), calibration (Negative Log-Likelihood), and rank correlation (Spearman \(\rho\)). As the task involves parameter estimation rather than text generation, encoder-only models suffice. Decoder architectures are unnecessary, as Chronocept operates at the application layer, interfacing with downstream systems without altering core language models.

\subsection{Baseline Models and Training Setup}
We benchmark Chronocept against a representative set of baselines spanning statistical (LR, SVR), tree-based (XGB), and neural architectures (FFNN, Bi-LSTM, BERT Regressor). Each baseline is trained to jointly predict \(\xi\), \(\omega\) and \(\alpha\) from BERT-based input embeddings of the parent text and temporal subtexts. Targets are Z-Score normalized to standardize learning across all models.

\begin{figure}[h]
\centering
\includegraphics[width=1\linewidth]{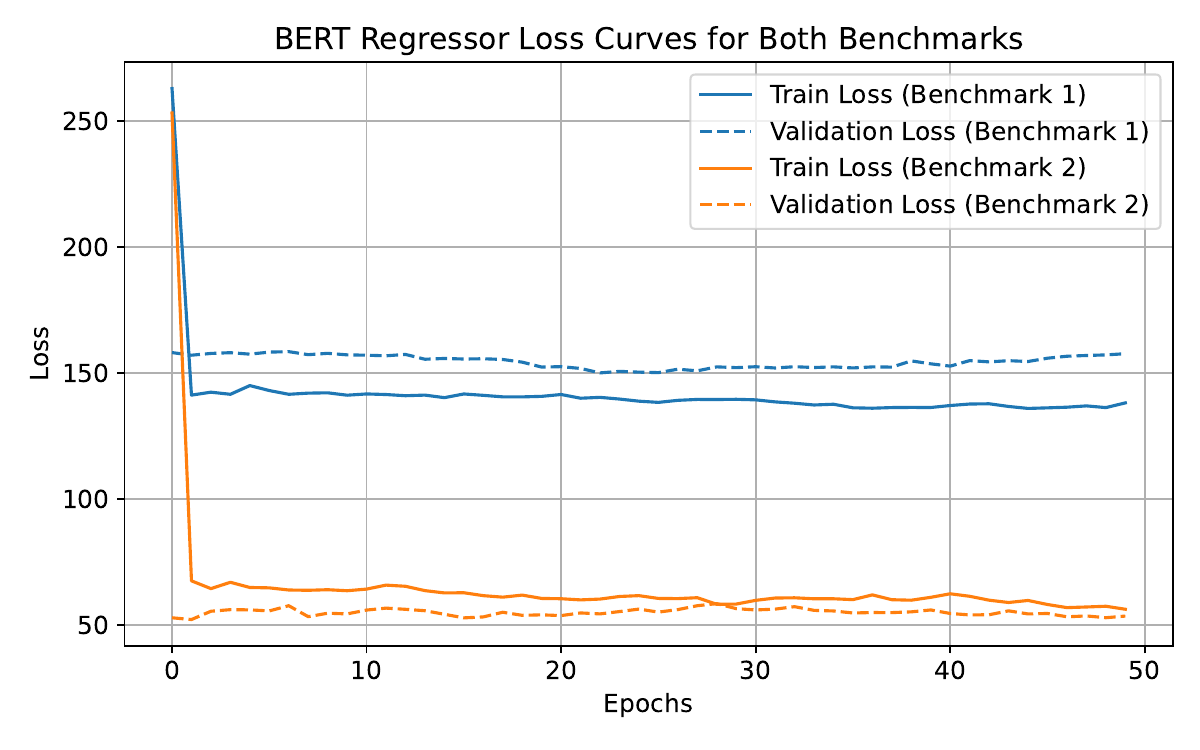}
\caption{BERT training loss curves for Benchmark I and Benchmark II. The loss flatlined after 2 epochs for both benchmarks.}
\label{fig:bert_regressor_loss}
\end{figure}

Hyperparameters for all baselines (except BERT) were tuned via grid search; final configurations are detailed in \autoref{appendix:hyperparameter_training}. FFNN and Bi-LSTM models were trained for 100 epochs while BERT was trained for 50 epochs. BERT training loss plateaued after approximately 2 epochs across both benchmarks, as shown in \autoref{fig:bert_regressor_loss}, suggesting early stopping could be beneficial for future experiments.

All training and inference experiments were conducted on a machine with an Intel Core i9-14900K CPU, 16GB DDR5 RAM, and an NVIDIA RTX 4060 GPU.

Baseline implementations and training scripts are publicly available at: \url{https://github.com/krishgoel/chronocept-baseline-models}.

\subsection{Quantitative Evaluation}
\autoref{tab:quantitative_results} summarizes the performance of baseline models across both benchmarks. Each reported metric reflects the mean score across the three predicted parameters.

Feedforward Neural Networks (FFNN) outperform all other models overall, achieving the lowest MSE, MAE, NLL, and the highest Spearman Correlation on Benchmark I. This supports prior findings that simpler architectures, when paired with high-quality pretrained embeddings, can match or exceed deeper models in accuracy and efficiency \citep{saphra-lopez-2019-understanding, wei2021pretrained}.

Bi-LSTM trails FFNN on Benchmark I but outperforms it on Benchmark II in four of five metrics - MSE, \(R^2\), NLL and Spearman \(\rho\) - on Benchmark II, which provides longer textual context. This is consistent with prior findings on sequence modeling \citep{meng-rumshisky-2018-context, dligach-etal-2017-neural}, and may stem from Bi-LSTM's ability to better model long-range dependencies, while FFNNs rely on the BERT [CLS] token, which can struggle to encode longer contexts into a single vector \citep{li-etal-2020-sentence}.

BERT Regression improves significantly from Benchmark I to II, with MSE dropping by over 50\%, suggesting longer inputs help stabilize fine-tuning. However, BERT still underperforms across all metrics, consistent with its known sensitivity to overfitting and gradient noise on small regression datasets \citep{mosbach2021stability, peters-etal-2019-tune, lee2020biobert}.

Among classical models, SVR and XGBoost perform reasonably but are outpaced by neural approaches. SVR achieves relatively strong \(R^2\) and NLL scores on Benchmark I, while XGBoost and LR lag across all metrics. Their interpretability and training efficiency still make them useful reference baselines \citep{drucker1996support, rogers-etal-2020-primer}.

Together, these results affirm that pretrained embeddings paired with compact neural regressors like FFNN yield state-of-the-art performance. Additionally, they highlight how models with sequence-awareness, such as Bi-LSTM and BERT, benefit disproportionately from longer contexts.

\begin{table*}[h]
    \centering
    \fontsize{10pt}{12pt}\selectfont
    \sisetup{group-separator={,}, round-mode=places, round-precision=3}
    \begin{tabular}{lcccccccccc}
        \toprule
        \textbf{Metric} & \multicolumn{2}{c}{\textbf{MSE}} & \multicolumn{2}{c}{\textbf{MAE}} & \multicolumn{2}{c}{\(\mathbf{R^2}\)} & \multicolumn{2}{c}{\textbf{NLL}} & \multicolumn{2}{c}{\textbf{Spearman}} \\
        \cmidrule(lr){2-3} \cmidrule(lr){4-5} \cmidrule(lr){6-7} \cmidrule(lr){8-9} \cmidrule(lr){10-11}
        \textbf{Baseline} & BI & BII & BI & BII & BI & BII & BI & BII & BI & BII \\
        \midrule
        LR         & 1.3610 & 1.1009 & 0.9179 & 0.8361 & -0.3610 & -0.1009 & 1.5730 & 1.4670 & 0.2338 & 0.3279 \\
        XGB        & 0.8884 & 0.9580 & 0.7424 & 0.8011 & 0.1116  & 0.0420  & 1.3598 & 1.3975 & 0.2940 & 0.2331 \\
        SVR        & 0.9067 & 0.8889 & 0.7529 & 0.7740 & 0.0933  & 0.1111  & 1.3700 & 1.3601 & 0.3281 & 0.3293 \\
        FFNN       & \textbf{0.8763} & 0.8715 & \textbf{0.7284} & \textbf{0.7583} & \textbf{0.1237}  & 0.1285  & \textbf{1.3529} & 1.3502 & \textbf{0.3543} & 0.3437 \\
        Bi-LSTM    & 0.9203 & \textbf{0.8702} & 0.7571 & 0.7646 & 0.0797  & \textbf{0.1298}  & 1.3774 & \textbf{1.3494} & 0.2367 & \textbf{0.3535} \\
        BERT       & 145.8611 & 68.1507 & 6.7570 & 4.6741 & -0.0090 & -0.1122 & 3.9103 & 3.5299 & -0.0485 & -0.2407 \\
        \bottomrule
    \end{tabular} 
    \caption{Test set performance of baseline models for Benchmark I (BI) and Benchmark II (BII). Lower values for MSE, MAE, and NLL indicate better performance; higher \(R^2\) and Spearman Correlation \(\rho\) denote improved fit.}
    \label{tab:quantitative_results}
\end{table*}

\subsection{Impact of Temporal Axes: Ablation Studies}
To assess the utility of explicit temporal axes in Chronocept, we conduct two ablation studies on Benchmark 1 using the Bi-LSTM and FFNN baselines.

The first study evaluates the impact of removing all axis-level information, and the second examines the impact of randomly shuffling axis order during training. This setup parallels prior work on robustness testing via perturbed input labels \citep{moradi-samwald-2021-evaluating}.

Both the axis‑removal and axis‑shuffle setups lead to substantial performance degradation, indicating that both - the presence and consistent ordering of temporal axes - play a key role in accurately modeling temporal validity.

\autoref{tab:ablation_summary} summarizes the increase in MSE for the Bi-LSTM baseline. Experimental design and complete results for both baselines are detailed in \autoref{appendix:multi_axis_study} (excluded axes) and \autoref{appendix:incorrect_axis_labels} (shuffled axes).

\begin{table}[h]
    \centering
    \begin{tabular}{lcc}
        \toprule
        \textbf{Ablation Type} & \textbf{Ablated MSE} & \textbf{Increase} \\
        \midrule
        Exclusion of Axes    & 0.9625 & 4.59\% \\
        Erroneous Labeling  & 1.0107 & 9.83\% \\
        \bottomrule
    \end{tabular}
    \caption{Ablation results for the Bi-LSTM baseline. Relative increases are computed over the original MSE of $0.9203$.}
    \label{tab:ablation_summary}
\end{table}

\section{Conclusion \& Applications}
\label{sec:conclusion_and_applications}
We introduced Chronocept, a framework that models temporal validity as a continuous probability distribution using a unified, parameterized representation. By encoding validity through location (\(\xi\)), scale (\(\omega\)), and skewness (\(\alpha\)), Chronocept provides a generalizable mathematical scheme for temporal reasoning in language.

Through structured annotations and explicit temporal axes, Chronocept enables models to capture not just if, but when and for how long information remains valid - advancing beyond binary truth labels to a richer temporal understanding.

Empirical results highlight the effectiveness of simple neural models paired with pretrained embeddings, and ablation studies underscore the importance of structural consistency and axis-level decomposition.

Chronocept opens pathways for temporally aware applications, including retrieval-augmented generation (RAG), fact verification, knowledge lifecycle modeling, and proactive AI agents that act based on temporal salience \citep{miksik2020building}. All datasets, annotations, and baselines are publicly released to support continued research in this space.

\section{Limitations}
\label{sec:limitations}
In this section, we highlight key limitations of Chronocept and suggest directions for future refinement and broader applicability.

\paragraph{Unimodal Temporal Representation.}
Chronocept models temporal validity with a unimodal, single-peaked distribution. While this ensures interpretability and efficient annotation, it cannot represent events with multiple distinct periods of relevance, such as seasonal or recurring phenomena.

\paragraph{Sentence-Level Context Only.}
The dataset consists of short, self-contained sentences without document-level or historical context. This limits the modeling of long-range temporal dependencies and evolving narratives, constraining discourse-level temporal reasoning.

\paragraph{No Atemporality Indicators.}
Chronocept lacks explicit labels for atemporal or universally valid facts, introducing ambiguity between permanently valid and time-sensitive information.

\paragraph{Minimum Validity Constraint from Log Time Scale.}
The logarithmic time scale imposes a lower bound of one minute, making it unsuitable for modeling events that become instantly obsolete, such as flash updates or ephemeral statements.

\section{Acknowledgments}
\label{sec:acknowledgments}
We thank Mohammed Iqbal, Meenakshi Kumar, Yudhajit Mondal, Tanish Sharma, Devansh Sharma, Lakshya Paliwal, Ishaan Verma, and Sanjit Chitturi for their help with data annotation.

\appendix
\section*{Appendix}

\section{Annotation Guidelines}
\label{appendix:annotation_guidelines} 
This section outlines the annotation guidelines used in the Chronocept dataset. These were introduced through an in-person training session and remained accessible throughout the annotation phase via a custom Streamlit-based interface for annotations\footnote{\url{https://streamlit.io}}. The guidelines provide precise instructions for temporal segmentation, axis categorization, and temporal validity distribution plotting, supplemented with definitions, examples, and coverage of edge cases for all eight temporal axes.

During the initial warm-up phase, annotators exhibited substantial confusion between the Generic and Static axes. To mitigate this, the guidelines were revised to incorporate clearer contextual definitions and axis-specific "key questions" designed to improve disambiguation. These revisions led to a marked improvement in inter annotator agreement.

\begin{figure*}[h]
    \centering
    \includegraphics[height=1\textheight]{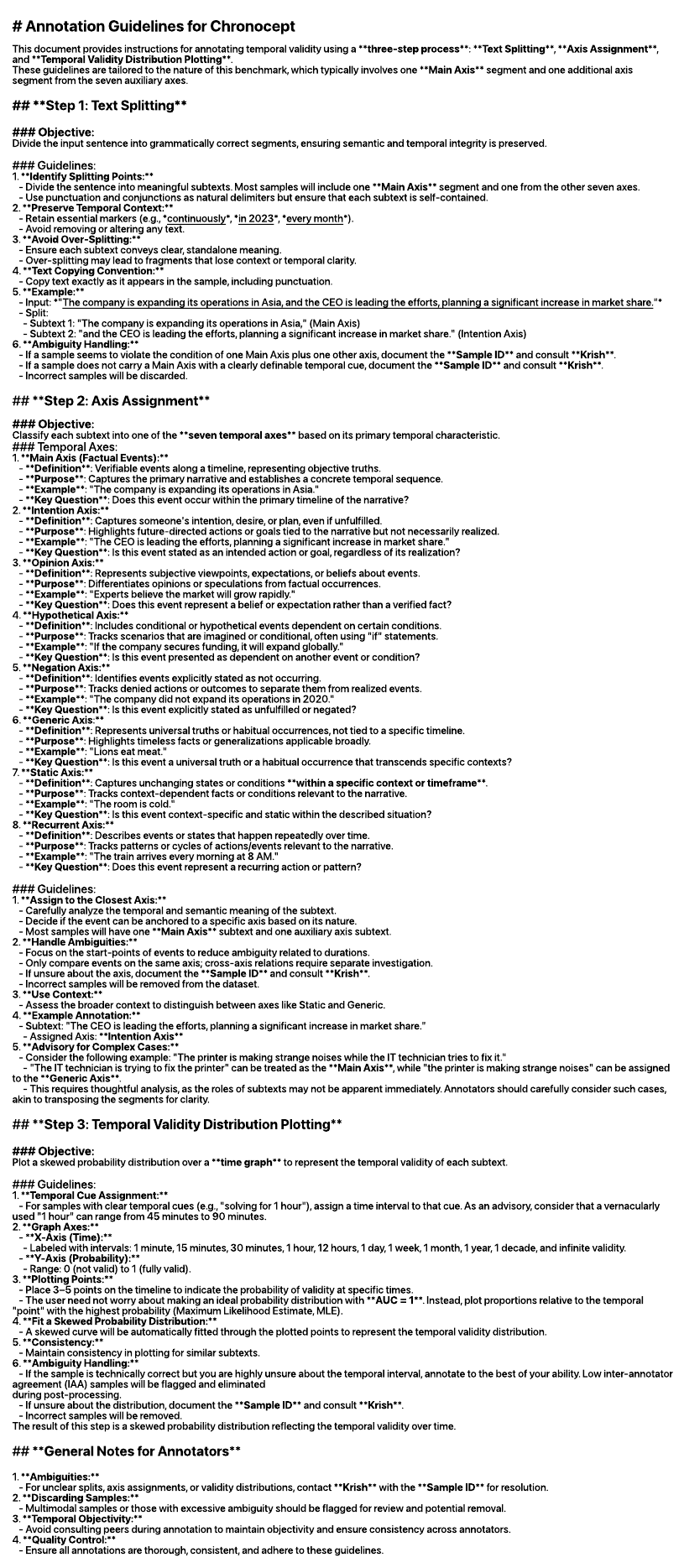}
    \caption{Annotation guidelines for Chronocept.}
    \label{fig:annotation_guidelines}
\end{figure*}

The complete guidelines are shown in \autoref{fig:annotation_guidelines}.

\section{Axis Confusion Analysis: Generic and Static}
\label{appendix:generic_static_confusion_analysis}
\begin{figure}[H]
    \centering
    \begin{subfigure}[b]{0.48\textwidth}
        \includegraphics[width=\textwidth]{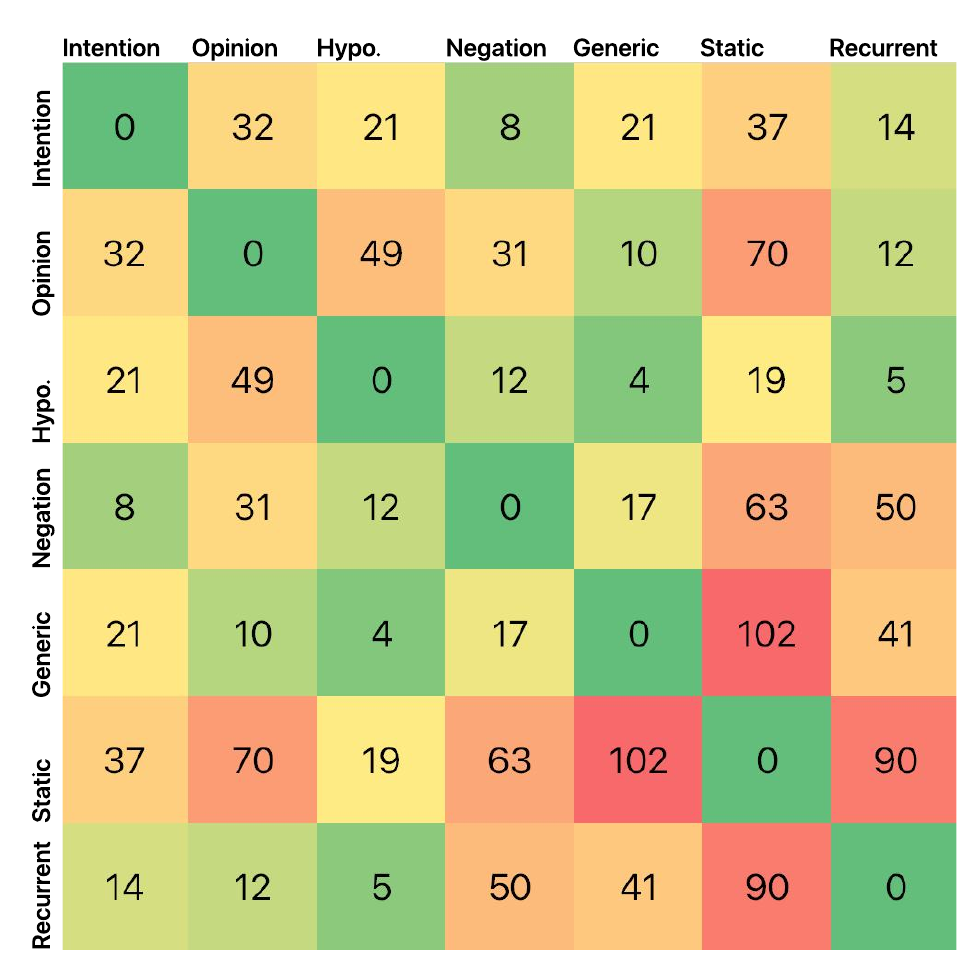}
        \caption{Axis assignment co-occurrence matrix with Generic and Static treated as distinct classes}
    \end{subfigure}\hfill
    \begin{subfigure}[b]{0.48\textwidth}
        \includegraphics[width=\textwidth]{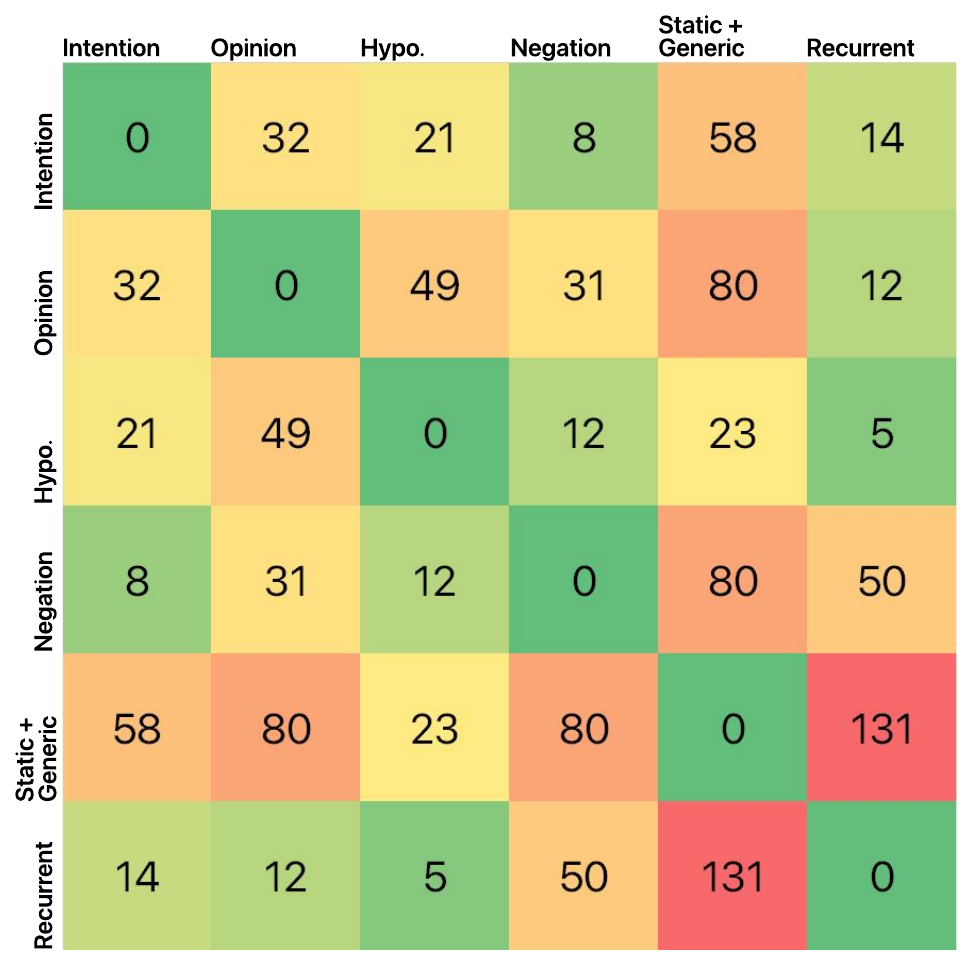}
        \caption{Axis assignment co-occurrence matrix after merging Generic and Static into a unified class}
    \end{subfigure}
    \caption{Comparison of co-occurrence matrices before and after merging the Generic and Static axes, used to assess annotation consistency.}
    \label{fig:confusion_matrices}
\end{figure}

This appendix investigates a key source of annotator disagreement in the Chronocept annotation process: the difficulty in consistently distinguishing between the Generic and Static temporal axes.

Generic segments typically express habitual or timeless statements, while Static segments describe ongoing but context-specific states. Their semantic similarity led to frequent disagreement in axis assignment.

To address this, the annotation guidelines were refined during the warm-up phase with axis-specific clarifications and diagnostic questions. The guideline clarification led to reduced confusion, as shown in the co-occurrence matrices in \autoref{fig:confusion_matrices}.

While co-occurrence matrices are traditionally used to analyze disagreement patterns between annotators, we treat them here as confusion matrices by including agreement counts along the diagonal, enabling standard metric computation.

To quantify the benefit of merging these axes, we computed micro-averaged inter-annotator precision. Treating this as a multi-class classification task, we additionally calculate Cohen's Kappa to assess inter-annotator agreement beyond chance. As shown in \autoref{tab:metrics_comparison}, merging resulted in a consistent improvement across both metrics: precision improved by 18.0\% and Cohen's Kappa by 17.47\%.

\begin{table}[h]
    \centering
    \begin{tabular}{lcc}
        \toprule
        \textbf{Axis Setting} & \textbf{Precision} & \textbf{Cohen's Kappa} \\
        \midrule
        Original& 0.4443 & 0.3291 \\
        Merged & 0.5243 & 0.3866 \\
        \bottomrule
    \end{tabular}
    \caption{Improvement in annotator alignment metrics after merging Generic and Static into a single class.}
    \label{tab:metrics_comparison}
\end{table}

\section{Time Scale Logarithm Base Conversion}
\label{appendix:log_conversion}
\begin{figure}[H]
    \centering
    \includegraphics[width=\linewidth]{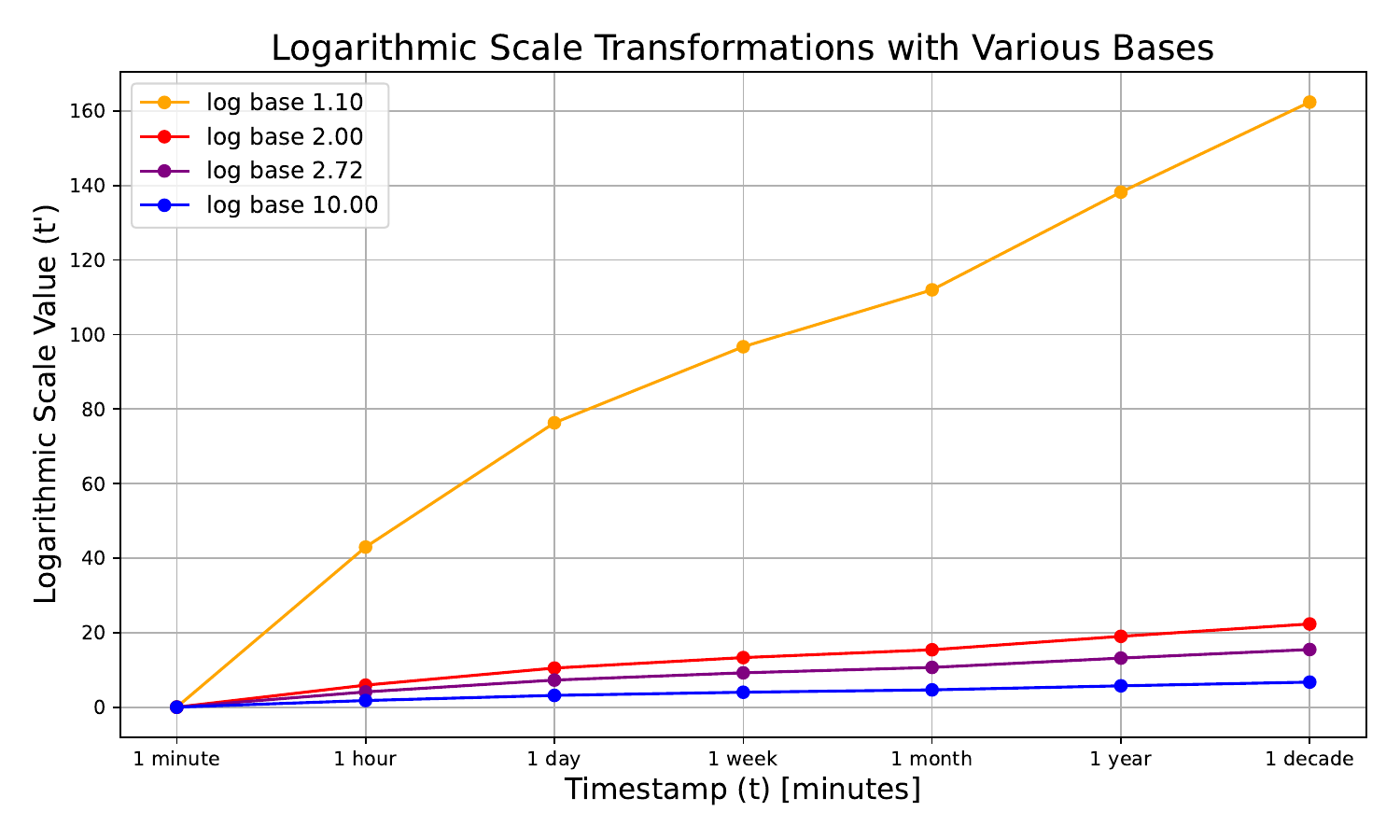}
    \caption{Effect of logarithmic base choice on time axis representation. Base 1.1 preserves quasi-linear spacing; larger bases induce stronger compression.}
    \label{fig:log_bases}
\end{figure}

Chronocept represents time on a logarithmic axis to unify short- and long-term temporal dynamics in a compact space. The transformation is defined over a configurable base \(b\); all released datasets use base \(1.1\). A reusable \texttt{DataLoader} with log conversion is available in the official baselines repository\footnote{\url{https://github.com/krishgoel/chronocept-baseline-models}}.

\paragraph{Log Transformation.}
Given time \(t\) in minutes, the log-space representation is:
\[
t' = \frac{\ln(t)}{\ln(b)}.
\]
Base \(1.1\) yields quasi-linear spacing across intervals like hours, days, and years, preserving interpretability. \autoref{fig:log_bases} shows that higher bases increasingly compress longer intervals, while base \(1.1\) maintains resolution across scales.

\paragraph{Compression Analysis.}
\autoref{tab:log_compression_analysis} summarizes the compression effect across bases \(1.1\), \(2\), and \(10\). For each timestamp, we report the log value \(t'\), compression ratio \(\text{CR} = t'/t\), and percentage compression.

\begin{table*}[h]
    \centering
    \fontsize{10}{12}\selectfont
    \begin{tabular}{l c c c c c c c c c c}
        \toprule
        & & \multicolumn{3}{c}{\textbf{log base 1.1}} & \multicolumn{3}{c}{\textbf{log base 2}} & \multicolumn{3}{c}{\textbf{log base 10}} \\
        \cmidrule(lr){3-5} \cmidrule(lr){6-8} \cmidrule(lr){9-11}
        \textbf{Timestamp} & \textbf{Linear (t)} & \textbf{\(t'\)} & \textbf{CR} & \textbf{\%} & \textbf{\(t'\)} & \textbf{CR} & \textbf{\%} & \textbf{\(t'\)} & \textbf{CR} & \textbf{\%} \\
        \midrule
        1 minute & 1 & 0.0 & 0.000 & 100 & 0.0 & 0.000 & 100 & 0.0 & 0.000 & 100 \\
        1 hour & 60 & 42.96 & 0.716 & 28.4 & 5.91 & 0.099 & 90.1 & 1.78 & 0.030 & 97.0 \\
        1 day & 1440 & 76.30 & 0.053 & 94.7 & 10.47 & 0.007 & 99.3 & 3.16 & 0.002 & 99.8 \\
        1 week & 10080 & 96.73 & 0.010 & 99.0 & 13.30 & 0.001 & 99.9 & 4.00 & 3.968e-4 & 99.9 \\
        1 month & 43200 & 111.97 & 0.003 & 99.7 & 15.39 & 3.563e-4 & 99.9 & 4.63 & 1.072e-4 & \verb|~|100 \\
        1 year & 525600 & 138.23 & 2.623e-4 & \verb|~|100 & 19.00 & 3.615e-5 & \verb|~|100 & 5.72 & 1.088e-5 & \verb|~|100 \\
        1 decade & 5256000 & 162.25 & 3.087e-5 & \verb|~|100 & 22.33 & 4.249e-6 & \verb|~|100 & 6.72 & 1.279e-6 & \verb|~|100 \\
        \bottomrule
    \end{tabular}
    \caption{Compression analysis across logarithmic bases. CR = \(t'/t\), Compression \% = \(100 \times (1 - \text{CR})\).}
    \label{tab:log_compression_analysis}
\end{table*}

To convert values between log bases \(m\) and \(b\):
\[
t'^{(b)} = \frac{\ln(m)}{\ln(b)} \cdot t'^{(m)}.
\]

\paragraph{Skew-Normal Parameter Adjustment.}
Chronocept models temporal validity using a skew-normal distribution:
\[
f(x;\,\xi,\omega,\alpha) = \frac{2}{\omega}\,\phi\left(\frac{x - \xi}{\omega}\right)\,\Phi\left(\alpha\,\frac{x - \xi}{\omega}\right),
\]
where \(\xi\) and \(\omega\) denote location and scale. When converting between bases:
\[
\xi^{(b)} = \frac{\ln(m)}{\ln(b)} \cdot \xi^{(m)}, \quad \omega^{(b)} = \frac{\ln(m)}{\ln(b)} \cdot \omega^{(m)}.
\]
Skewness \(\alpha\) remains invariant.

\section{Comparison of Distributions for Modeling Temporal Validity and Curve Fitting Methodology}
\label{appendix:curve_fitting}
This section evaluates candidate distributions for modeling temporal validity and outlines the curve fitting methodology. We consider six synthetic, unimodal scenarios varying along three axes: \textit{offset} (peak position), \textit{duration} (span of validity), and \textit{asymmetry} (skew in rise and decay). \autoref{tab:temporal_scenarios} lists a representative sentence and five annotation points per scenario, placed on a base-1.1 logarithmic time axis.

Each temporal profile is defined by a smooth freehand curve from which five points are sampled—one at the peak, two mid-validity, and two low-validity points. These define a proportional shape used for fitting.

Since these curves represent relative probabilities, their area under the curve (AUC) is unconstrained. During optimization, a scaling factor is applied to fit freely, followed by Trapezoidal Rule normalization to enforce AUC = 1 while preserving shape.

To reduce computational overhead over long-tailed domains, we recommend rescaling the fitted curve by its maximum value to constrain it to \([0, 1]\). This avoids instability from very small values in AUC-normalized densities. The result, while no longer a true probability distribution, retains shape and relative comparisons. We refer to it as a \textit{proportional validity curve}, useful in applications prioritizing ranking or visualization over strict probabilistic semantics.\\

\noindent
Candidate distributions include:\\

\noindent
\textbf{Gaussian Normal:}
\[
f_{Gaussian}(x; \mu, \sigma) = \frac{1}{\sqrt{2\pi}\,\sigma}\exp\!\left(-\frac{(x-\mu)^2}{2\sigma^2}\right)
\]

\noindent
\textbf{Exponential:}
\[
f_{Exp}(x; \lambda) = \lambda\exp(-\lambda x), \text{where }x \ge 0
\]

\noindent
\textbf{Log-normal:}
\[
f_{LN}(x; \mu, \sigma) = \frac{1}{x\sqrt{2\pi}\,\sigma}\exp\!\left(-\frac{(\ln x - \mu)^2}{2\sigma^2}\right),
\]
\[
\text{where }x > 0
\]

\noindent
\textbf{Gamma:}
\[
f_{\Gamma}(x; k, \theta) = \frac{1}{\Gamma(k)\,\theta^k}x^{\,k-1}\exp\!\left(-\frac{x}{\theta}\right),
\]
\[
\text{where }x > 0
\]

\noindent
\textbf{Skewed Normal:}
\[
f_{SN}(x; \xi, \omega, \alpha) = \frac{2}{\omega}\,\phi\!\left(\frac{x-\xi}{\omega}\right)\,\Phi\!\left(\alpha\,\frac{x-\xi}{\omega}\right)
\]
where \(\phi(z)\) is the standard normal PDF and \(\Phi(z)\) is the standard normal CDF.\\

\noindent\textbf{Optimization:} Parameter estimation is performed using the Trust Region Reflective (TRF) algorithm by minimizing the sum of squared residuals:
\[
SSR(\theta) = \sum_{i=1}^{N}\left(y_i - f(x_i;\theta)\right)^2
\]
This is implemented via \texttt{scipy.optimize.curve\_fit}\footnote{\url{https://docs.scipy.org/doc/scipy/reference/generated/scipy.optimize.curve_fit.html}}. After optimization, we compute:
\[
N = \int_{x_{\min}}^{x_{\max}} f_{\text{fit}}(x)\,dx,
\]
\[
f_{\text{norm}}(x) = \frac{f_{\text{fit}}(x)}{N},
\quad
f_{\max} = \max_x f_{\text{norm}}(x),
\]\[
S_{\text{final}} = \frac{S_{\text{fit}}}{N \cdot f_{\max}}
\]

\noindent\textbf{Evaluation:} RMSE is used as the primary goodness-of-fit metric. As a scale-sensitive measure that penalizes large deviations, a lower RMSE indicates superior fit quality.

\begin{figure*}[h]
    \centering
    \begin{subfigure}[b]{0.49\textwidth}
        \includegraphics[width=\textwidth]{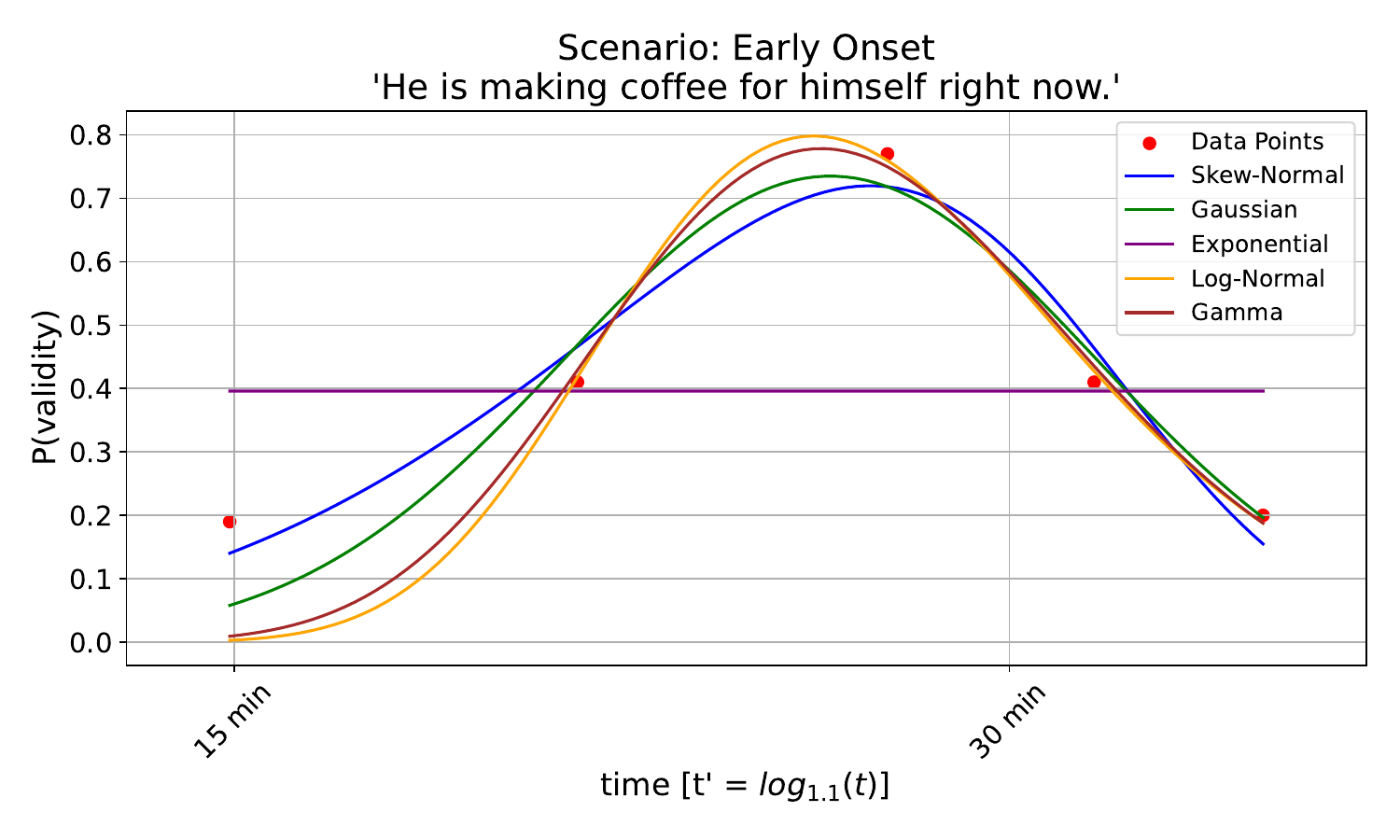}
        \caption{Early Onset: Peak validity occurs soon after publication.}
    \end{subfigure}
    \hfill
    \begin{subfigure}[b]{0.49\textwidth}
        \includegraphics[width=\textwidth]{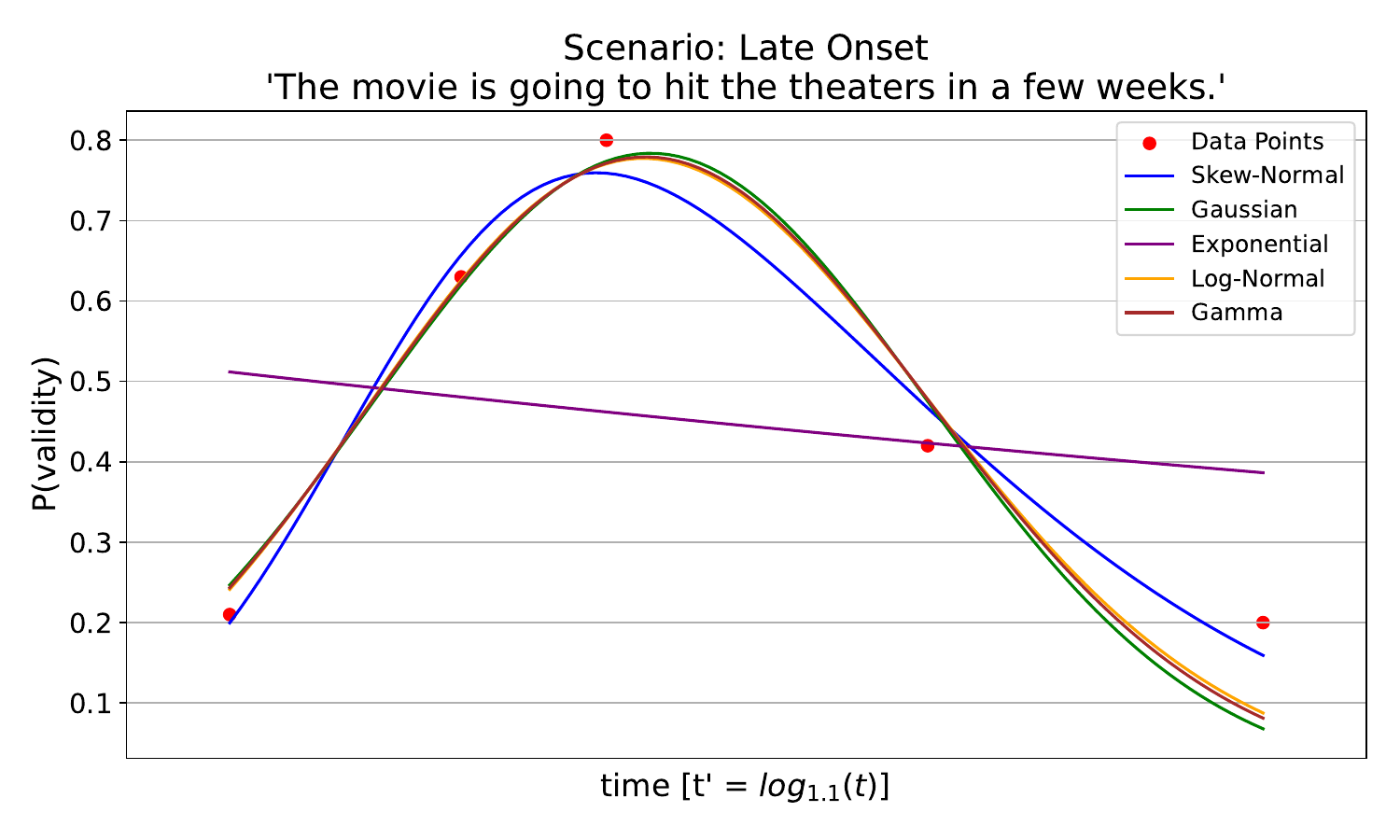}
        \caption{Late Onset: Validity emerges gradually and peaks later.}
    \end{subfigure}

    \begin{subfigure}[b]{0.49\textwidth}
        \includegraphics[width=\textwidth]{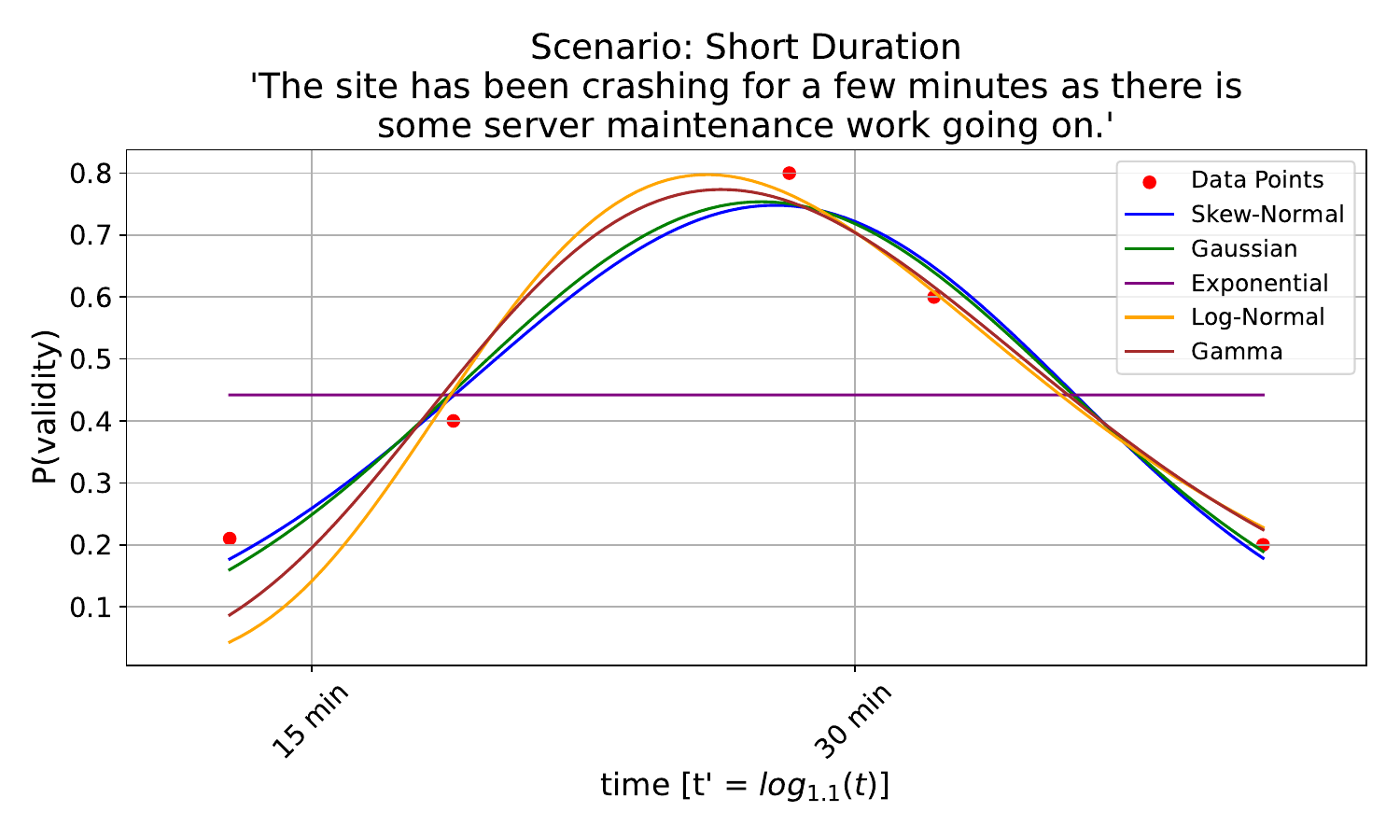}
        \caption{Short Duration: A narrow window of high relevance.}
    \end{subfigure}
    \hfill
    \begin{subfigure}[b]{0.49\textwidth}
        \includegraphics[width=\textwidth]{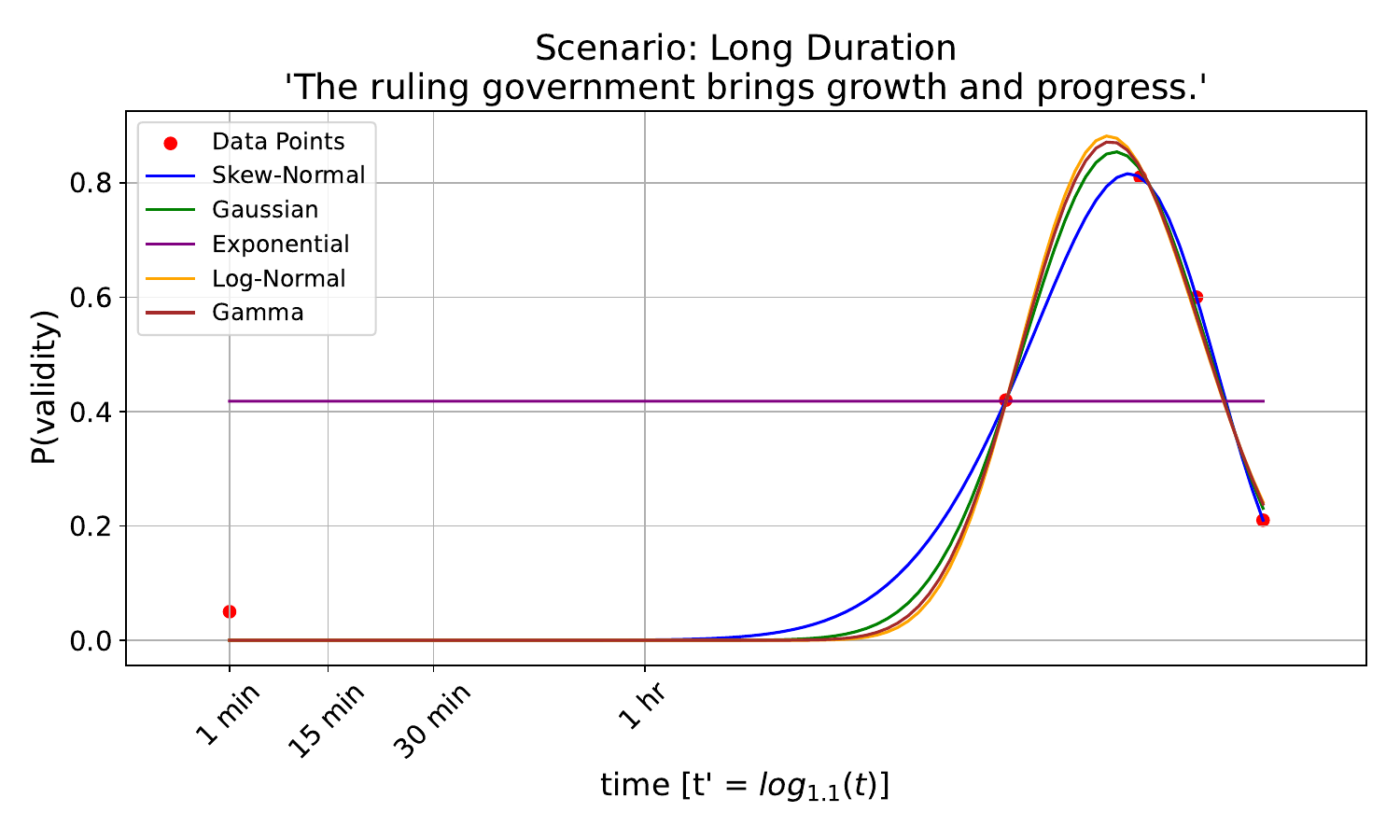}
        \caption{Long Duration: Validity persists over time.}
    \end{subfigure}

    \begin{subfigure}[b]{0.49\textwidth}
        \includegraphics[width=\textwidth]{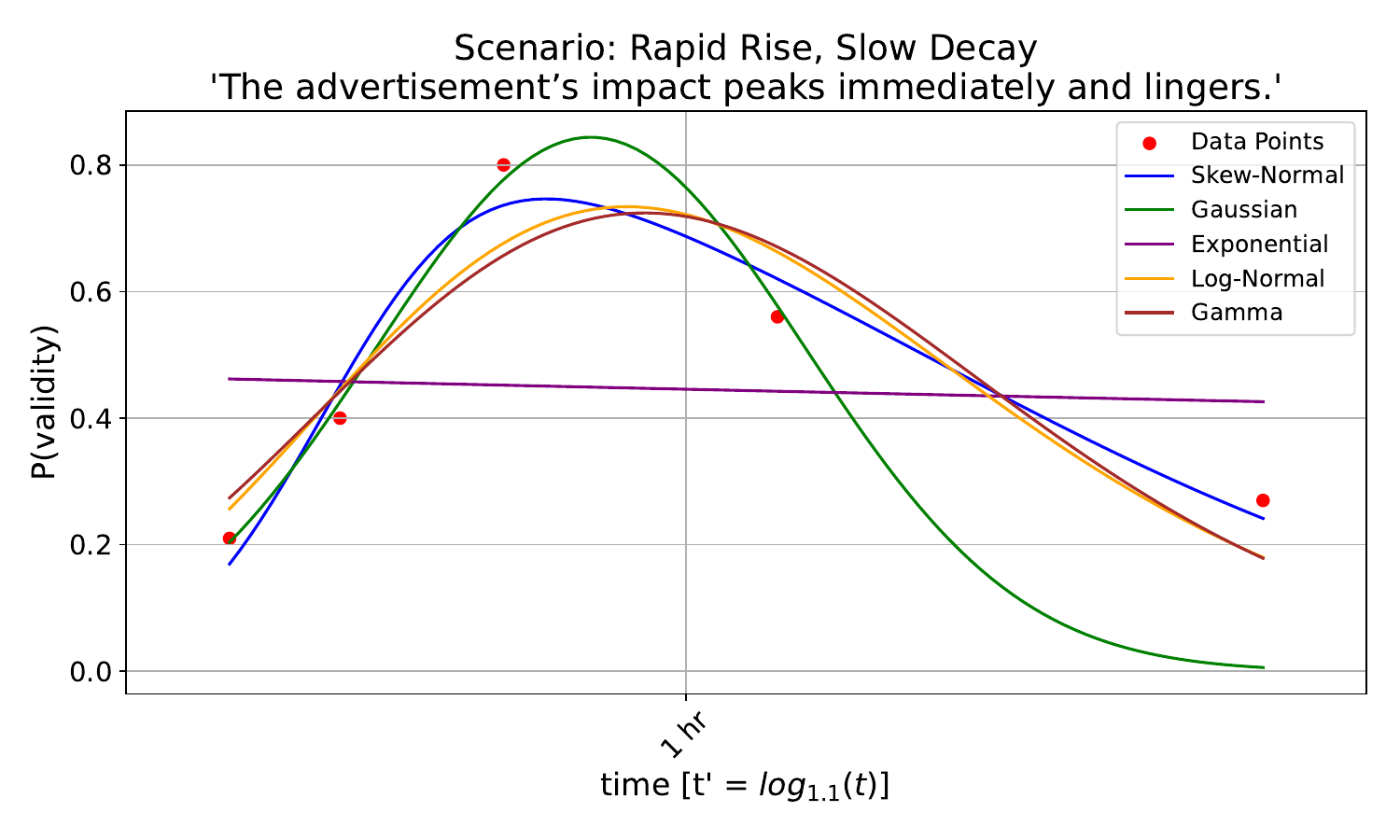}
        \caption{Rapid Rise, Slow Decay: Sudden onset, gradual decline.}
    \end{subfigure}
    \hfill
    \begin{subfigure}[b]{0.49\textwidth}
        \includegraphics[width=\textwidth]{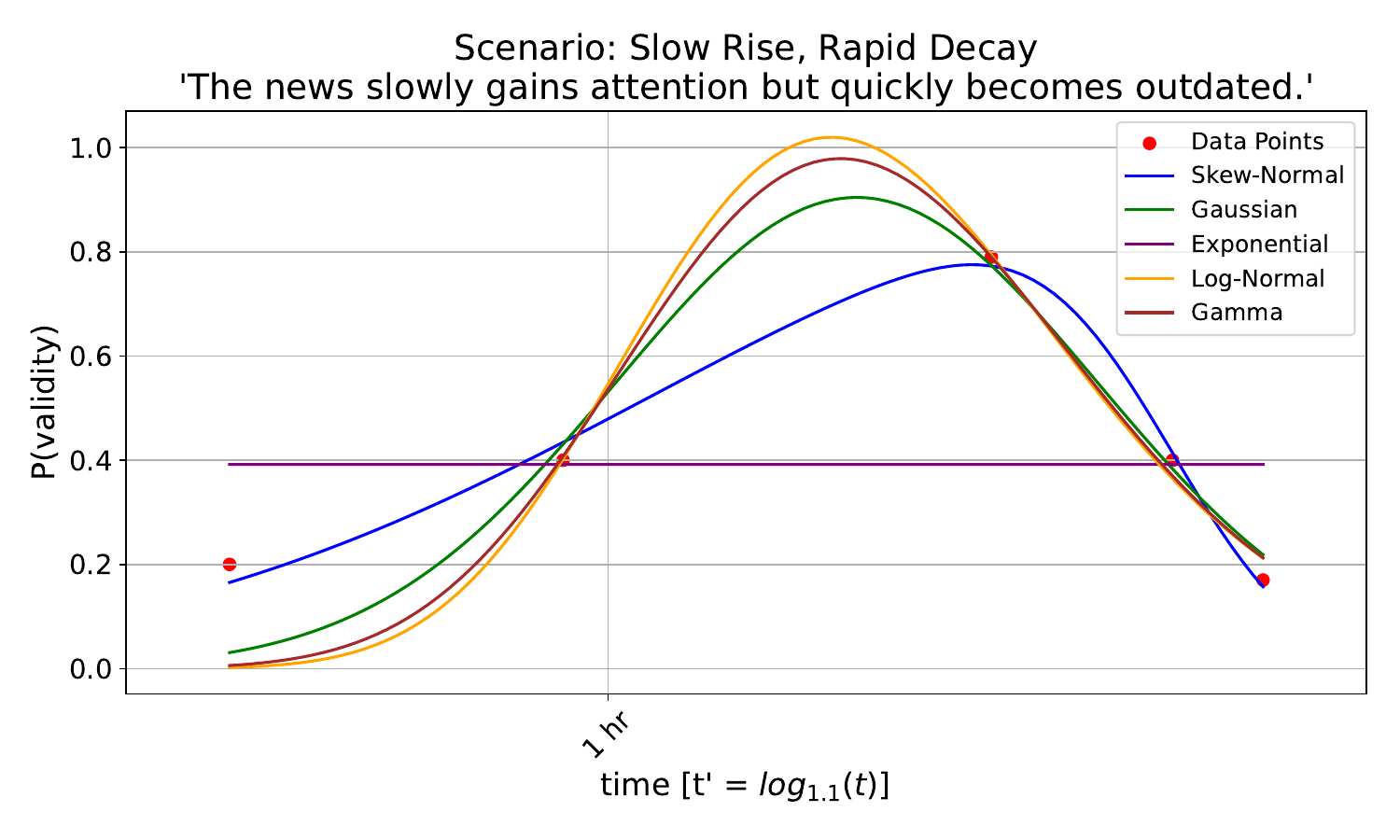}
        \caption{Slow Rise, Rapid Decay: Gradual onset, sharp drop.}
    \end{subfigure}
    
    \caption{Visual fit comparison of candidate distributions across six temporal scenarios. The skew-normal consistently provides the best fit, modeling varied validity patterns in onset, duration, and asymmetry.}
    \label{fig:temporal_scenarios_plots}
\end{figure*}

\autoref{tab:temporal_scenarios} and \autoref{fig:temporal_scenarios_plots} present the six scenarios, annotation points, and corresponding fitted curves. \autoref{tab:curve_fitting_comparison} reports RMSE for each candidate distribution across scenarios. The skew-normal consistently yields the lowest RMSE, confirming its suitability for modeling asymmetric and variable-duration temporal profiles.

\begin{table*}[h]
    \centering
    \renewcommand{\arraystretch}{1.3}
    \begin{tabularx}{\textwidth}{>{\raggedright\arraybackslash}X 
                                 >{\raggedright\arraybackslash}X 
                                 >{\raggedright\arraybackslash}X}
        \toprule
        \textbf{Temporal Scenario} & \textbf{Sample Sentence} & \textbf{Annotation Points \((x, y)\)} \\
        \midrule
        S1: Early Onset & 
        "He is making coffee for himself right now." & 
        (14.91, 0.19), (21.64, 0.41), (27.64, 0.77), (31.64, 0.41), (34.91, 0.20) \\

        S2: Late Onset & 
        "The movie is going to hit the theaters in a few weeks." & 
        (93.75, 0.21), (100.67, 0.80), (106.57, 0.42), (112.73, 0.20), (98.0, 0.63) \\

        S3: Short Duration & 
        "The site has been crashing for a few minutes as there is some server maintenance work going on." & 
        (12.73, 0.21), (28.19, 0.80), (41.28, 0.20), (32.19, 0.60), (18.91, 0.40) \\

        S4: Long Duration & 
        "The ruling government brings growth and progress." & 
        (1, 0.05), (130.38, 0.81), (147.84, 0.21), (111.29, 0.42), (138.38, 0.60) \\

        S5: Rapid Rise, Slow Decay & 
        "The advertisement's impact peaks immediately and lingers." & 
        (42.73, 0.21), (46.91, 0.40), (53.10, 0.80), (63.46, 0.56), (81.83, 0.27) \\

        S6: Slow Rise, Rapid Decay & 
        "The news slowly gains attention but quickly becomes outdated." & 
        (43.28, 0.20), (58.01, 0.40), (76.92, 0.79), (84.92, 0.40), (88.92, 0.17) \\
        \bottomrule
    \end{tabularx}
    \caption{Six temporal scenarios illustrating the effects of offset, duration, and asymmetry. Each scenario is represented by 5 annotation points on a log-transformed time axis with base $1.1$.}
    \label{tab:temporal_scenarios}
\end{table*}

\begin{table*}[h]
    \centering
    \begin{tabular}{lcccccccl}
        \toprule
        \textbf{Distribution} & \textbf{S1} & \textbf{S2} & \textbf{S3} & \textbf{S4} & \textbf{S5} & \textbf{S6} & \textbf{Parameters} \\
        \midrule
        Gaussian      & 0.0709 & 0.0673 & 0.0424 & 0.0273 & 0.1193 & 0.0806 & $(\mu,\ \sigma)$ \\
        Exponential   & 0.2103 & 0.2291 & 0.2312 & 0.2704 & 0.2126 & 0.2212 & $(\lambda)$ \\
        Log-normal    & 0.0844 & 0.0597 & 0.0804 & 0.0325 & 0.0872 & 0.0919 & $(\mu,\ \sigma)$ \\
        Gamma         & 0.0827 & 0.0623 & 0.0668 & 0.0307 & 0.0968 & 0.0899 & $(k,\ \theta)$ \\
        Skewed Normal & \textbf{0.0514} & \textbf{0.0357} & \textbf{0.0407} & \textbf{0.0224} & \textbf{0.0505} & \textbf{0.0247} & $(\xi,\ \omega,\ \alpha)$ \\
        \bottomrule
    \end{tabular}
    \caption{Average RMSE values for candidate distributions across six temporal scenarios. All distributions were fitted using a scaling factor \(S\) to enforce AUC \(=1\). A lower RMSE indicates a better fit, as RMSE heavily penalizes large errors due to squaring, is scale-dependent, and more sensitive to outliers.}
    \label{tab:curve_fitting_comparison}
\end{table*}

\section{Synthetic Generation of Samples}
\label{appendix:synthetic}
This section presents the plaintext markdown prompts used for synthetic dataset generation in Chronocept via the GPT-o1 model \citep{openai2024o1}. The prompts are designed to yield syntactically coherent text with explicit temporal structure. Generation was performed in batches of 50 samples per prompt.

\begin{figure*}[h]
    \centering
    \includegraphics[width=1\linewidth]{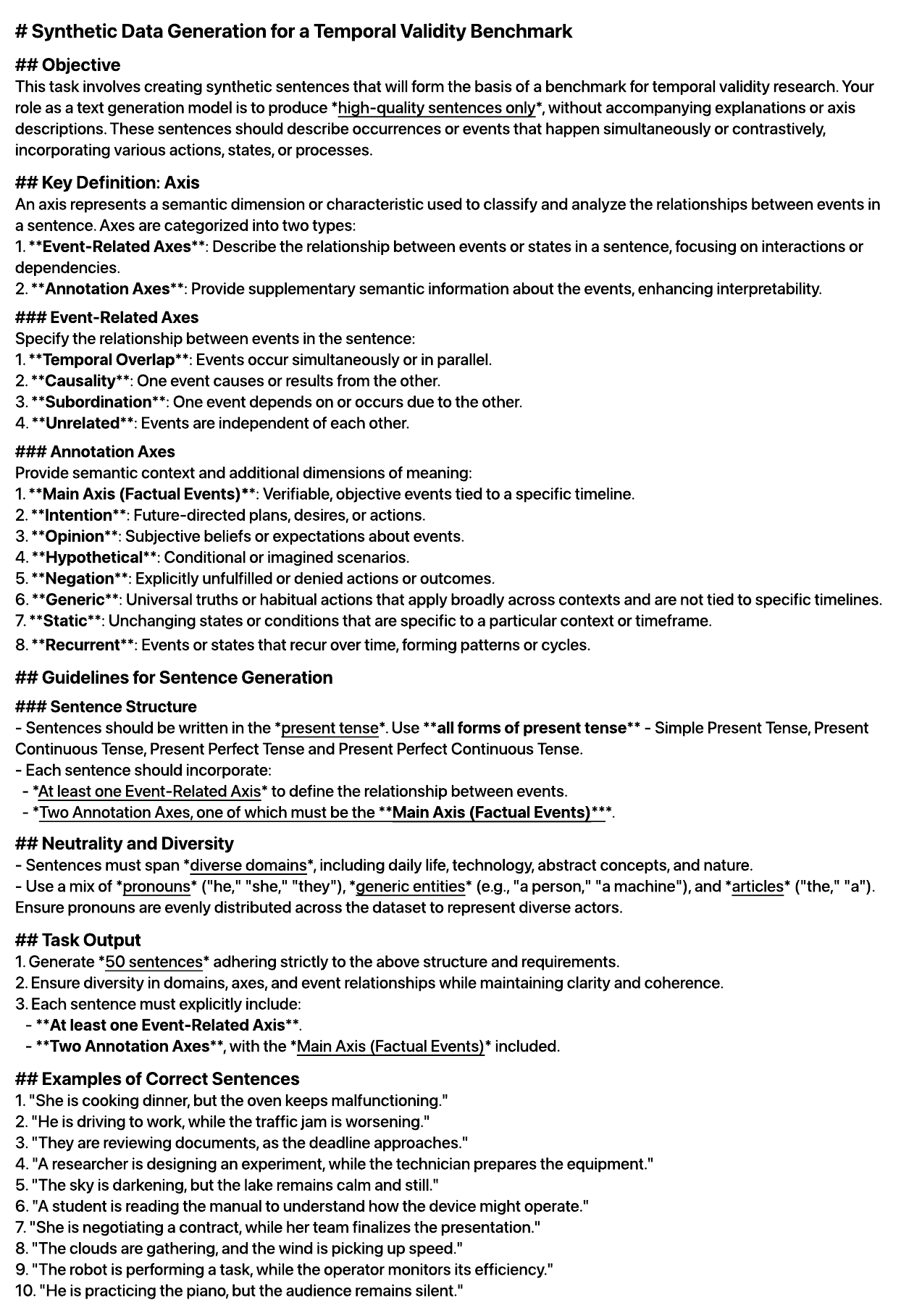}
    \caption{Plaintext markdown prompt for Benchmark I.}
    \label{fig:prompt_benchmark_I}
\end{figure*}

\begin{figure*}[h]
    \centering
    \includegraphics[width=1\linewidth]{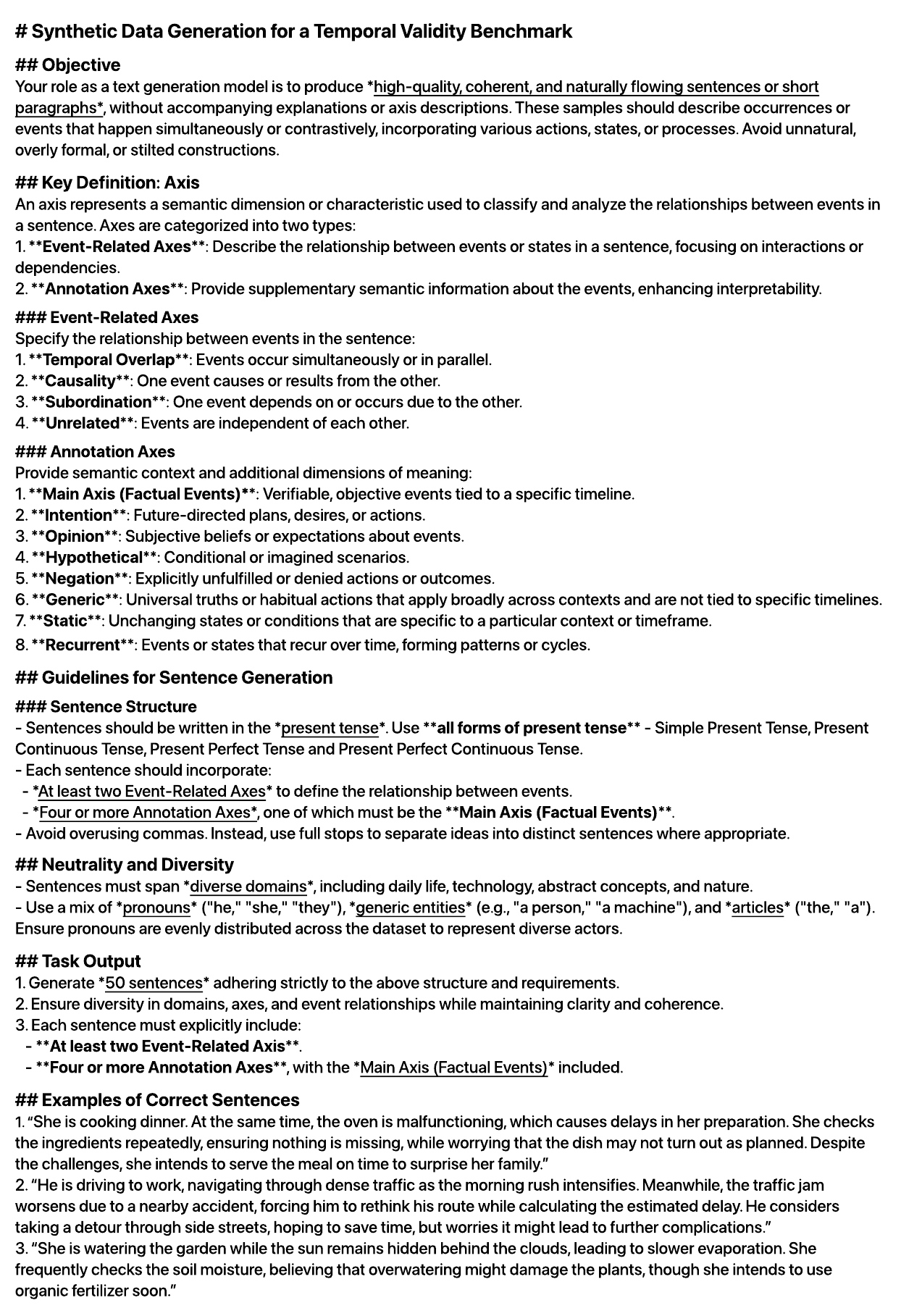}
    \caption{Plaintext markdown prompt for Benchmark II.}
    \label{fig:prompt_benchmark_II}
\end{figure*}

The prompts are shown in \autoref{fig:prompt_benchmark_I} for Benchmark-I and \autoref{fig:prompt_benchmark_II} for Benchmark-II.

\section{Ablation Study: Impact of Structured Temporal Axes on Model Performance}
\label{appendix:multi_axis_study}
To evaluate the contribution of multi-axis temporal annotations in modeling temporal validity, we conduct an ablation study on the Bi-LSTM and FFNN baselines. Specifically, we assess the effect of removing structured temporal axes from the model input.

\paragraph{Input Construction.}
Each example in Chronocept is annotated along multiple temporal axes. In the standard setup, axis-specific embeddings are concatenated in a fixed order to the embedding of the parent text, forming a structured input representation. The ablation removes these axis embeddings, retaining only the parent text embedding.

\paragraph{Setup.}
We compare the two configurations (with and without axis embeddings) using Bi-LSTM and FFNN models on Benchmark I. Both models are trained to predict the parameters \(\xi\), \(\omega\), and \(\alpha\) of the skew-normal temporal validity distribution. Evaluation is performed using MSE, MAE, \(R^2\), NLL, and CRPS.

\paragraph{Results.}
\autoref{tab:axis_inclusion_ablation} reports the results for both models. Including axis embeddings reduces Bi‑LSTM MSE by 4.6\% and boosts \(R^2\) by 112\%, confirming that structured cues matter more for goodness‑of‑fit than for absolute error. FFNN sees a 6.9\% MSE drop and a 95.7\% gain in \(R^2\), exhibiting a similar trend with even greater error reduction across all metrics.

These findings are consistent with prior work showing that compositional and auxiliary structure improves model generalization and fit across tasks \citep{lake2018generalizationsystematicitycompositionalskills, sogaardanders2016lowlevelsupervised}.

\begin{table*}[h]
    \centering
    \fontsize{10}{12}\selectfont
    \begin{tabular}{lcccccc}
        \toprule
        \textbf{Model} & \textbf{Setting} & \textbf{MSE} & \textbf{MAE} &
        \textbf{\(R^2\)} & \textbf{NLL} & \textbf{CRPS} \\
        \midrule
        \multirow{3}{*}{Bi‑LSTM}
        & Without Axes          & 0.9625 & 0.7659 & 0.0375 & 1.3998 & 0.7659 \\
        & Absolute Change (\(\Delta\)) & 0.0422 & 0.0088 & 0.0422 & 0.0224 & 0.0088 \\
        \cmidrule(lr){2-7}
        & \textit{Improvement}  & 4.59\% & 1.16\% & 112.53\% & 1.63\% & 1.16\% \\
        \midrule
        \multirow{3}{*}{FFNN}
        & Without Axes          & 0.9368 & 0.7531 & 0.0632 & 1.3863 & 0.7531 \\
        & Absolute Change (\(\Delta\)) & 0.0605 & 0.0247 & 0.0605 & 0.0334 & 0.0247 \\
        \cmidrule(lr){2-7}
        & \textit{Improvement}  & 6.91\% & 3.39\% & 95.71\%  & 2.47\% & 3.39\% \\
        \bottomrule
    \end{tabular}
    \caption{Ablation results on Benchmark I for Bi-LSTM and FFNN with axis embeddings removed. “Absolute Change” rows show differences from the original metrics in \autoref{tab:quantitative_results}.}
    \label{tab:axis_inclusion_ablation}
\end{table*}

\paragraph{Conclusion.}
Structured axis embeddings improve performance across both architectures, particularly in \(R^2\), which nearly doubles, indicating better distributional alignment. These results validate Chronocept’s use of explicit temporal structure and are consistent with prior work on structured auxiliary signals.

\section{Ablation Study: Impact of Incorrect Temporal Axes Labeling}
\label{appendix:incorrect_axis_labels}
We evaluate the sensitivity of temporal validity modeling to erroneous axis labelling by conducting an ablation on FFNN and Bi-LSTM baselines. Specifically, we shuffle the order of temporal axis embeddings during training while preserving correct ordering in the test set.

\paragraph{Setup.}
In Chronocept, input representations are formed by concatenating temporal axis embeddings in a fixed sequence with the parent text embedding. This ablation introduces erroneous axis labelling by disrupting the axis order during training, thereby breaking the structural alignment. The evaluation set remains unperturbed. Models are trained to predict skew-normal parameters \(\xi\), \(\omega\), and \(\alpha\), and evaluated on Benchmark I using MSE, MAE, \(R^2\), NLL, and CRPS.

\paragraph{Results.}  
\autoref{tab:ablation_misalignment} shows that misaligned axis ordering during training degrades performance significantly. Bi-LSTM MSE increases by 9.81\% and \(R^2\) decreases by 113.43\%; FFNN sees a 13.36\% MSE increase and 94.58\% \(R^2\) decrease. These results suggest that disrupting structural alignment introduces inductive noise, echoing prior findings on the role of compositional structure \citep{lake2018generalizationsystematicitycompositionalskills} and input robustness \citep{moradi-samwald-2021-evaluating}. The pronounced drop in \(R^2\) highlights that axis ordering is critical for fit quality.    

\begin{table*}[h]
    \centering
    \fontsize{10}{12}\selectfont
    \begin{tabular}{lcccccc}
        \toprule
        \textbf{Model} & \textbf{Setting} & \textbf{MSE} & \textbf{MAE} & \textbf{\(R^2\)} & \textbf{NLL} & \textbf{CRPS} \\
        \midrule
        \multirow{3}{*}{Bi‑LSTM}
        & Erroneous Axes        & 1.0107 & 0.7984 & -0.0107 & 1.4243 & 0.7984 \\
        & Absolute Change (\(\Delta\)) & 0.0904 & 0.0413 & \(-0.0904\) & 0.0469 & 0.0413 \\
        \cmidrule(lr){2-7}
        & \textit{Performance Drop} & 9.81\% & 5.46\% & 113.43\% & 3.40\% & 5.46\% \\
        \midrule
        \multirow{3}{*}{FFNN}
        & Erroneous Axes        & 0.9933 & 0.7591 & 0.0067  & 1.4156 & 0.7591 \\
        & Absolute Change (\(\Delta\)) & 0.1170 & 0.0307 & \(-0.1170\) & 0.0627 & 0.0307 \\
        \cmidrule(lr){2-7}
        & \textit{Performance Drop} & 13.36\% & 4.21\% & 94.58\% & 4.63\% & 4.21\% \\
        \bottomrule
    \end{tabular}
    \caption{Ablation results on Benchmark I for Bi-LSTM and FFNN under erroneous temporal axis labelling during training. “Absolute Change” rows show differences from the original metrics in \autoref{tab:quantitative_results}.}    
    \label{tab:ablation_misalignment}
\end{table*}

\paragraph{Conclusion.}
Erroneous axis labelling during training leads to statistically significant drops in performance, particularly in \(R^2\), highlighting the importance of Chronocept’s structured multi-axis representation for accurate temporal modeling.

\section{Hyperparameter Search and Final Baseline Configurations}
\label{appendix:hyperparameter_training}
All baseline models were tuned via grid search on the validation split of each benchmark. All neural models except BERT were trained for 100 epochs, with early stopping applied based on validation loss when applicable. BERT was trained for 50 epochs. Final hyperparameters are summarized below.

\paragraph{Support Vector Regression (SVR).}
We searched over \(C \in \{0.1, 1, 10\}\), \(\varepsilon \in \{0.01, 0.1, 1\}\), and kernel type \(\in \{\textit{linear}, \textit{rbf}\}\). The optimal setting used an RBF kernel with \(C = 1\) and \(\varepsilon = 1\) (see \autoref{tab:svr_final_hyperparameters}).

\begin{table}[H]
    \centering
    \begin{tabular}{lccc}
        \toprule
        \textbf{Benchmark} & \(C\) & \(\varepsilon\) & \textbf{Kernel} \\
        \midrule
        Benchmark I & 1 & 1 & rbf \\
        Benchmark II & 1 & 1 & rbf \\
        \bottomrule
    \end{tabular}
    \caption{Final SVR hyperparameters.}
    \label{tab:svr_final_hyperparameters}
\end{table}

\paragraph{Linear Regression (LR).}
The grid search over \texttt{fit\_intercept} \(\in \{\textit{True}, \textit{False}\}\) selected \textit{False} in both cases (see \autoref{tab:lr_final_hyperparameters}).

\begin{table}[H]
    \centering
    \begin{tabular}{lc}
        \toprule
        \textbf{Benchmark} & \textbf{Fit Intercept} \\
        \midrule
        Benchmark I & False \\
        Benchmark II & False \\
        \bottomrule
    \end{tabular}
    \caption{Final Linear Regression setting.}
    \label{tab:lr_final_hyperparameters}
\end{table}

\paragraph{XGBoost (XGB).}
We tuned \(n\_estimators \in \{50, 100\}\), \(max\_depth \in \{3, 5\}\), and learning rate \(\in \{0.1, 0.01\}\). The best configuration used 50 estimators, depth 3, and learning rate 0.1 (see \autoref{tab:xgb_final_hyperparameters}).

\begin{table}[H]
    \centering
    \begin{tabular}{lccc}
        \toprule
        \textbf{Benchmark} & \textbf{n} & \textbf{Depth} & \textbf{Learning Rate} \\
        \midrule
        Benchmark I & 50 & 3 & 0.1 \\
        Benchmark II & 50 & 3 & 0.1 \\
        \bottomrule
    \end{tabular}
    \caption{Final XGBoost hyperparameters.}
    \label{tab:xgb_final_hyperparameters}
\end{table}

\paragraph{Feedforward Neural Network (FFNN).}
We searched over hidden size \(\in \{64, 128, 256\}\), dropout \(\in \{0.0, 0.2, 0.5\}\), learning rate \(\in \{0.01, 0.001, 0.0001\}\), L1 regularization \(\in \{0.0, 0.0001, 0.001\}\), and weight decay \(\in \{0.0, 0.001, 0.01\}\). Final settings differed between benchmarks (see \autoref{tab:ffnn_final_hyperparameters}).

\begin{table}[H]
    \centering
    \begin{tabular}{lcc}
        \toprule
        \textbf{Benchmark} & \textbf{Hidden Dim} & \textbf{Learning Rate} \\
        \midrule
        Benchmark I & 64 & 0.001 \\
        Benchmark II & 256 & 0.01 \\
        \bottomrule
    \end{tabular}
    \caption{Final FFNN hyperparameters. Other parameters were fixed at: dropout = 0.0, L1 = 0.001, weight decay = 0.0.}
    \label{tab:ffnn_final_hyperparameters}
\end{table}

\paragraph{Bidirectional LSTM (Bi-LSTM).}
Search space included hidden size \(\in \{64, 128, 256\}\) and learning rate \(\in \{0.01, 0.001, 0.0001\}\). The final configuration used hidden size 64 and learning rate 0.0001 (see \autoref{tab:bilstm_final_hyperparameters}).

\begin{table}[H]
    \centering
    \begin{tabular}{lcc}
        \toprule
        \textbf{Benchmark} & \textbf{Hidden Dim} & \textbf{Learning Rate} \\
        \midrule
        Benchmark I & 64 & 0.0001 \\
        Benchmark II & 64 & 0.0001 \\
        \bottomrule
    \end{tabular}
    \caption{Final Bi-LSTM hyperparameters.}
    \label{tab:bilstm_final_hyperparameters}
\end{table}

\paragraph{BERT Regression.}
We tuned dropout \(\in \{0.0, 0.2, 0.4\}\) and learning rate \(\in \{0.0001\}\). The best setting used no dropout and learning rate 0.0001. Training loss converged within 2 epochs on both benchmarks (see \autoref{fig:bert_regressor_loss}).

\medskip

All scripts used for hyperparameter search and training are available at: \url{https://github.com/krishgoel/chronocept-baseline-models}.

\end{document}